\documentclass[10pt,twocolumn,letterpaper]{article}

\usepackage{wacv}              % To produce the CAMERA-READY version
\usepackage{graphicx}
\usepackage{amsmath}
\usepackage{amssymb}
\usepackage{booktabs}
\usepackage{multirow}
\usepackage{textcomp}
\usepackage{gensymb}
\usepackage{cite}
\usepackage{colortbl}
\usepackage{tikz}
\usepackage{pgfplots}
\usepackage{stfloats}

\usepackage{algorithm}
\usepackage{listings}

\usepackage{paralist}

\usepackage{xcolor}
\definecolor{rblue}{rgb}{0,0.5,1}
\definecolor{hollywoodcerise}{rgb}{0.96, 0.0, 0.63}
\definecolor{lasallegreen}{rgb}{0.03, 0.47, 0.19}
\definecolor{hanpurple}{rgb}{0.32, 0.09, 0.98}
\definecolor{green(pigment)}{rgb}{0.0, 0.65, 0.31}
\usepackage[pagebackref,breaklinks=true,colorlinks,bookmarks=false]{hyperref}
\hypersetup{colorlinks,linkcolor={red},citecolor={hanpurple},urlcolor={magenta}}

\usepackage[capitalize]{cleveref}
\crefname{section}{Sec.}{Secs.}
\Crefname{section}{Section}{Sections}
\Crefname{table}{Table}{Tables}
\crefname{table}{Tab.}{Tabs.}

\def\wacvPaperID{261} % *** Enter the WACV Paper ID here
\def\confName{WACV}
\def\confYear{2025}

\newcommand{\YZH}[1]{\textcolor{red}{#1}}

\pgfplotsset{compat=1.18}

\makeatletter
\renewcommand*{\@fnsymbol}[1]{\ensuremath{\ifcase#1\or *\or \dagger\or \ddagger\or
    \mathsection\or \mathparagraph\or \|\or **\or \dagger\dagger
    \or \ddagger\ddagger \else\@ctrerr\fi}}
\makeatother

\begin{document}

\title{EI-Nexus: Towards Unmediated and Flexible Inter-Modality Local Feature Extraction and Matching for Event-Image Data}

\author{Zhonghua Yi$^{1,2}$,
~~Hao Shi$^{1,2}$,
~~Qi Jiang$^{1,2}$,
~~Kailun Yang$^{3,}$\thanks{Corresponding authors (e-mail: {\tt kailun.yang@hnu.edu.cn, wangkaiwei@zju.edu.cn}).},
~~Ze Wang$^1$,\\
~~Diyang Gu$^1$,
~~Yufan Zhang$^1$,
~~Kaiwei Wang$^{1,2,*}$\\
\normalsize
$^1$State Key Laboratory of Extreme Photonics and Instrumentation, Zhejiang University\\
\normalsize
$^2$Jiaxing Research Institute, Zhejiang University
\normalsize
~~$^3$School of Robotics, Hunan University
}
\maketitle

%%%%%%%%% ABSTRACT
\begin{abstract}
Event cameras, with high temporal resolution and high dynamic range, have limited research on the inter-modality local feature extraction and matching of event-image data. We propose EI-Nexus, an unmediated and flexible framework that integrates two modality-specific keypoint extractors and a feature matcher. To achieve keypoint extraction across viewpoint and modality changes, we bring \textbf{L}ocal \textbf{F}eature \textbf{D}istillation (LFD), which transfers the viewpoint consistency from a well-learned image extractor to the event extractor, ensuring robust feature correspondence. Furthermore, with the help of \textbf{C}ontext \textbf{A}ggregation (CA), a remarkable enhancement is observed in feature matching. We further establish the first two inter-modality feature matching benchmarks, MVSEC-RPE and EC-RPE, to assess relative pose estimation on event-image data. Our approach outperforms traditional methods that rely on explicit modal transformation, offering more unmediated and adaptable feature extraction and matching, achieving better keypoint similarity and state-of-the-art results on the MVSEC-RPE and EC-RPE benchmarks. The source code and benchmarks will be made publicly available at \href{https://github.com/ZhonghuaYi/EI-Nexus_official}{EI-Nexus}.
\end{abstract}

%%%%%%%%% BODY TEXT
\section{Introduction}

\begin{figure}[!t]
    \centering
    \includegraphics[width=\linewidth]{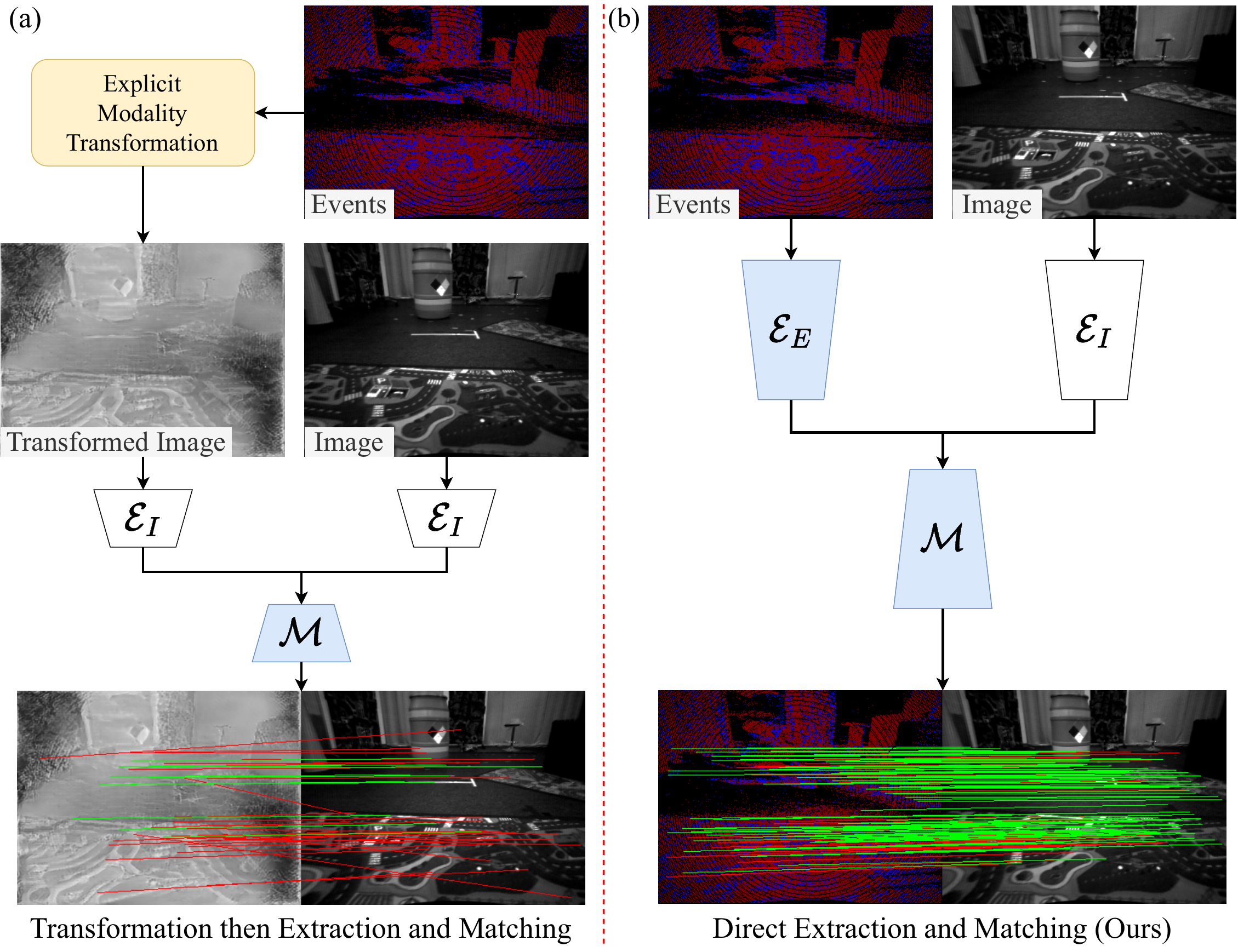}
    \vskip -0.7\baselineskip plus -1fil
    \caption{\textbf{Pipeline comparison.} (a) Traditional pipelines~\cite{muglikar2021calibrate, jiao2023lce} apply explicit modality transformation first, then utilize image-based extraction and matching models. (b) Our framework directly extracts keypoints from events and images and then applies feature matching, which is simpler and more powerful.}
    \label{fig:pipeline_comparison}
    \vskip -1.6\baselineskip plus -1fil
\end{figure}

Local feature extraction and feature matching are two related computer vision problems, with various downstream applications in SLAM~\cite{mur2015orb, wang2023lf, wang2024lf}, SfM~\cite{schonberger2016structure}, visual localization~\cite{sarlin2019coarse, bellavia2023progressive, shi2023panovpr}, tracking~\cite{zhou2020temporal}, and so on.
Local feature extraction~\cite{trajkovic1998fast, lowe2004distinctive, rublee2011orb} aims to extract sparse local features (\textit{i.e.} keypoints) from a given data, which shows invariant properties to environment changes.
Feature matching is supposed to find the correspondence between a pair of data.
Although correspondence could be obtained by many other techniques such as optical flow~\cite{yi2023focusflow, shi2023panoflow, shi2022csflow} and detector-free methods~\cite{sun2021loftr, wang2022matchformer}, the widely-used approaches~\cite{sarlin2020superglue, chen2021learning, lindenberger2023lightglue} are detector-based, by combing the local feature extraction into a whole matching procedure.

Feature extraction and matching have also been scaled to other modalities~\cite{mueggler2017fast, vasco2016fast, ramesh2019dart, li2019fa} like events.
Event camera as a new potential sensor that provides high temporal resolution and high dynamic range data, has been developed for many high-speed computer vision tasks, like motion deblur~\cite{sun2022event}, object tracking~\cite{messikommer2023data}, human pose estimation~\cite{chen2022efficient, yin2023rethinking}, and optical flow estimation~\cite{ye2023towards}. 
However, due to the modality difference, the event camera system is hard to cooperate with the RGB sensor system, especially on \emph{inter-modality} feature matching applications like calibration and registration.
Even though a few works~\cite{muglikar2021calibrate, jiao2023lce} attempt to overcome it by using explicit modality transform method~\cite{stoffregen2020reducing, scheerlinck2020fast} to translate the event data into video data and then use the image-oriented toolbox as illustrated in Fig.~\ref{fig:pipeline_comparison} (a), they are considered suboptimal due to the inherent information loss from sensor discrepancies.
As a result, exploring a more generalized inter-modality local feature extraction and matching method is urgent and important.

\begin{figure}[!t]
    \centering
    \includegraphics[width=\linewidth]{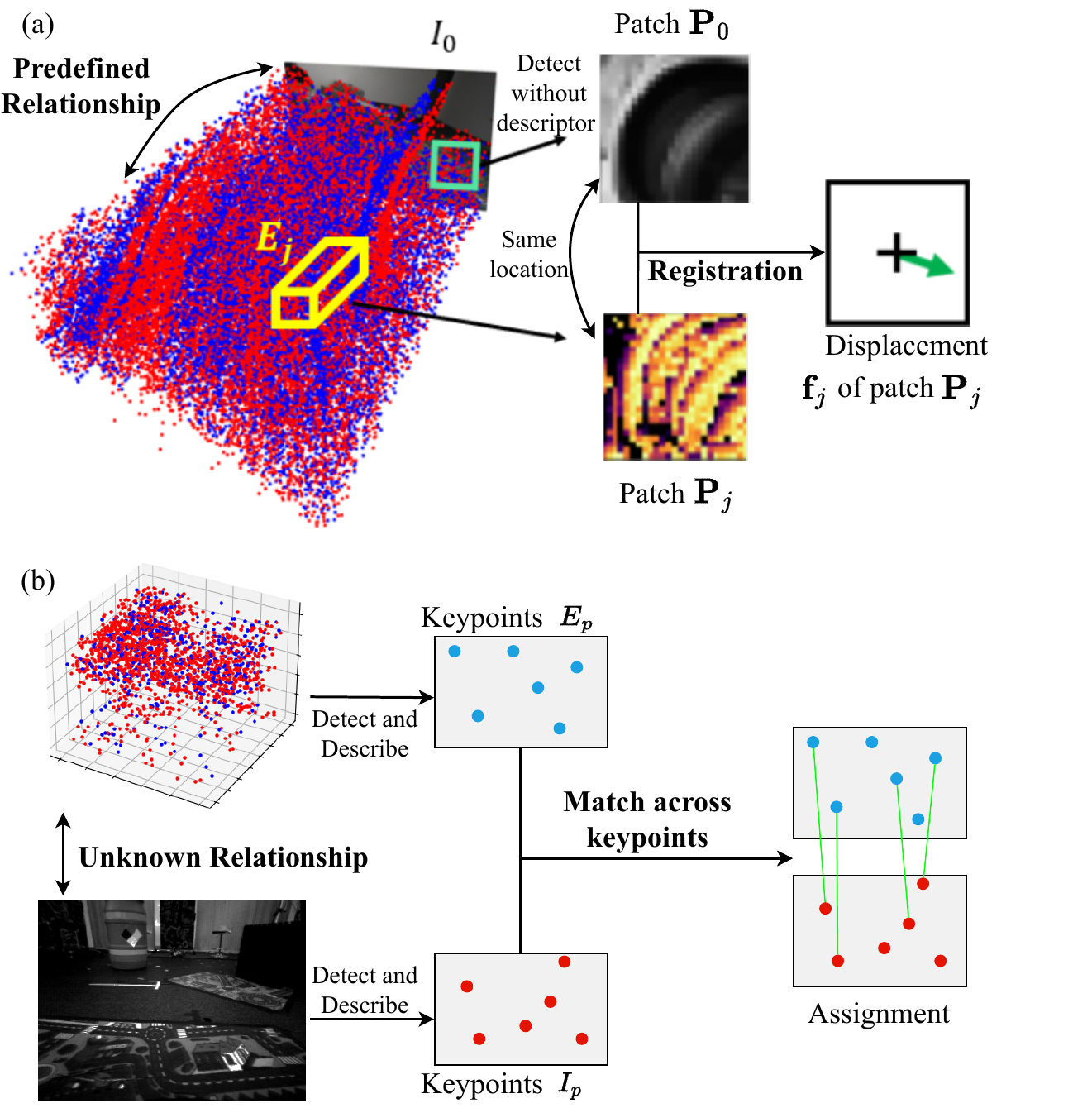}
    \vskip -0.7\baselineskip plus -1fil
    \caption{\textbf{Differences between (a) tracking using events and the reference frame and (b) inter-modality matching:} (a) Event-based feature tracking methods estimate displacement $\mathbf{f}_j$ of event patch $\mathbf{P}_j$ at $t_j$ from original image patch $\mathbf{P}_0$ at $t_0$ that locate in the same location. (b) Our inter-modality matching method separately extracts keypoints and descriptors and matches them using cross-modality descriptors, without the predefined relationship between events and images.}
    \label{fig:compare_tracking_matching}
    \vskip -1.6\baselineskip plus -1fil
\end{figure}

Recently, some event-based feature tracking methods~\cite{gehrig2018asynchronous, gehrig2020eklt, messikommer2023data} apply both events and frames for tracking, in which frames are treated as additional information to help tracking.
As shown in Fig.~\ref{fig:compare_tracking_matching}, they rely on the aligned image and events at timestamp $t_0$ to estimate displacement and do not provide robust descriptors, whereas our predict descriptors for matching without the need for a predefined relationship between events and image, making it applicable to a wide range of applications.

In this paper, we present a detector-based framework called EI-Nexus for unmediated and flexible \emph{inter-modality} local feature extraction and matching for event-image data.
Instead of learning the local feature extractor in a self-supervised way, we propose a novel local feature distillation (LFD) method to bring the viewpoint-invariant property of a well-learned image extractor into the event extractor.
To compensate for the intrinsic difference between different modalities, a learnable matching method with context aggregation is further introduced.
As shown in Fig.~\ref{fig:pipeline_comparison}, compared with traditional local feature extraction and matching pipeline, our framework reduces computation cost and performs much more powerfully, by matching keypoints directly extracted from different modalities.
The entire framework is flexible and each component is replaceable so it would benefit from any advanced image-oriented local feature extraction and matching approach.

Since no benchmark approach has been established to evaluate the performance of event-image inter-modality feature matching, two relative pose estimation benchmarks, MVSEC-RPE based on MVSEC dataset~\cite{zhu2018mvsec} and EC-RPE based on EC~\cite{mueggler2017event} are proposed, which offers novel challenges for further research.
We evaluate EI-Nexus on keypoint similarity and relative pose estimation tasks on them.
The experiments show that our framework achieves state-of-the-art keypoint similarity and relative pose estimation performance.
Furthermore, we test different image extractors, matching methods, and event representations, all of which significantly outperform traditional pipelines, demonstrating the flexibility of our framework.
At a glance, the contributions of this study are summarized as follows:
\begin{compactitem}
\item We are the first that introduce unmediated event-image \emph{inter-modality} local feature extraction and matching task for event-image data.

\item We design an unmediated and flexible detector-based framework for both local feature extraction and matching.

\item A novel local feature distillation is performed to train the event extractor during local feature extraction, and context aggregation shows as a better choice for inter-modality feature matching. 

\item Two relative pose estimation benchmarks MVSEC-RPE and EC-RPE are firstly established for the task of event-image \emph{inter-modality} local feature extraction and matching.
\end{compactitem}

\section{Related Work}
\noindent\textbf{Local Feature Extraction.}
Local feature extraction focuses on extracting several sparse keypoints with descriptors on an imaging plane, which are robust to the environment changes such as viewpoint and illumination. 
Early works~\cite{harris1988combined, trajkovic1998fast, lowe2004distinctive, rublee2011orb} utilize hand-crafted approaches to extract keypoints and their descriptors on a given image but suffer inconsistency from large environment changes.
Recent learning-based keypoint extraction techniques~\cite{detone2018superpoint, tyszkiewicz2020disk, gleize2023silk} could exhibit much more robustness. 
SuperPoint~\cite{detone2018superpoint} employs a self-supervised manner to extract keypoint locations and descriptors at the pixel level jointly.
SiLK~\cite{gleize2023silk} proposes a probabilistic approach and achieves better performance in many settings.

With the increasing interest and adoption of event cameras, keypoint extraction methods for event cameras are also being developed.
EvFast~\cite{mueggler2017fast} employs FAST~\cite{trajkovic1998fast} to detect interest points via timestamp difference. 
EvHarris~\cite{vasco2016fast} selects local features by transforming the event stream to a Time-Surface (TS) representation and then applying Harris detector~\cite{harris1988combined}.
Early approaches~\cite{mueggler2017fast, vasco2016fast, alzugaray2018asynchronous, lagorce2014asynchronous, clady2017motion, li2019fa, lagorce2015spatiotemporal} do not provide descriptors until DART~\cite{ramesh2019dart} calculates a log-polar-based feature descriptor robust to
scale, rotation, and viewpoint variations.
Learning-based methods like random forest~\cite{manderscheid2019speed} and CNNs~\cite{huang2023eventpoint, chiberre2021detecting} are also developed to extract keypoints from event data.
However, to the best of the authors' knowledge, a method for extracting cross-domain local features from image and event data with modality differences is scarce in the literature.
EI-Nexus addresses this problem by building on two independent local feature extractors to provide modality-invariant local features for multi-modal applications.

\noindent\textbf{Local Feature Matching.}
The goal of feature matching is to estimate an assignment between the local features of a data pair.
After keypoint extraction, early works employ methods such as the ratio test~\cite{lowe2004distinctive}, mutual check, or other filtering techniques~\cite{yi2018learning} to obtain robust matching results.
However, traditional matching methods degrade substantially in situations with significant noise caused by environmental changes or the sensor itself.
Learning methods~\cite{sarlin2020superglue, chen2021learning, lindenberger2023lightglue} are proposed to solve the problem.
Through Context Aggregation (CA), which aims to refine or transform descriptors from a pair of images before matching, these models achieve better performance in feature matching tasks like homography estimation~\cite{balntas2017hpatches} and relative pose estimation~\cite{li2018megadepth}.
Meanwhile, although detector-free approaches achieve more precise matching results~\cite{sun2021loftr, wang2022matchformer}, they cannot extract sparse local features from a single frame and incur higher computational costs.
Given the capacity of CA in implicitly transforming descriptors, it is suitable to serve as a descriptor translator for different modalities. 
Building on this, a flexible matching module is proposed in EI-Nexus which not only supports traditional matching methods but also incorporates learnable CA approaches.

\section{Methodology}

\begin{figure*}[!t]
    \centering
    \includegraphics[width=\linewidth]{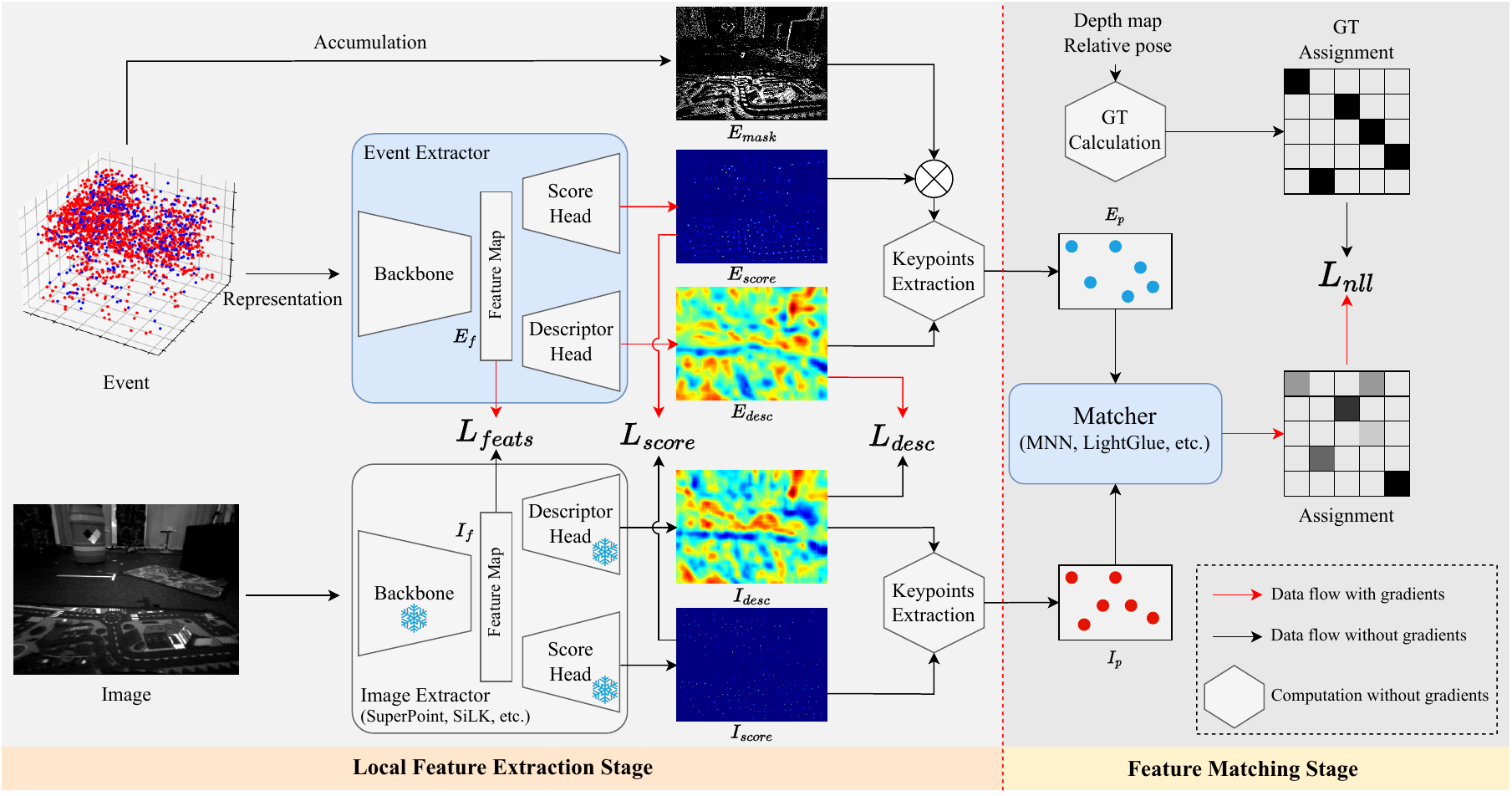}
    \caption{\textbf{Framework overview.} 
    The snowflakes represent no parameter optimization during training.
    Our framework follows a detector-based architecture including a local feature extraction stage and a feature matching stage. The event and image are separately sent to extractors to obtain the corresponding score map and descriptor map. Then the same keypoints extraction procedure is adopted for two branches, resulting in two keypoint sets. The two keypoint sets are sent to the matcher for feature matching, then the assignment matrix is finally estimated. During training, the event extractor is first trained through Local Feature Distillation (LFD), and then the matcher is trained through the ground-truth assignment calculated from the depth map and relative pose. Every component in the framework is modular, showing the flexibility of our design.}
    \label{fig:framework}
    \vskip -1.0\baselineskip plus -1fil
\end{figure*}

\subsection{Problem Formulation}
\label{sec:problem formulation}

\noindent\textbf{Intra-Modality Local Feature Extraction.}
Given a scene $\mathcal{L}(x,y,z,t)$, data $\mathcal{D}$ is captured from an observation $\mathcal{O}$.
A local feature extraction method $\mathcal{E}$ aims to predict a sparse keypoint set $\mathcal{D}_p$ containing positions $\mathbf{p}_i$ and descriptors $\mathbf{d}_i$ of all keypoints $i$, where the latter is invariant to $\mathcal{O}$:
\begin{equation}
\mathcal{D}_p= \{ (\mathbf{p}_i, \mathbf{d}_i) \} = \mathcal{E}(\mathcal{D}).
\label{eq:1}
\end{equation}

\noindent\textbf{Intra-Modality Feature Matching.}
Given a data pair $\{ \mathcal{D}^1, \mathcal{D}^2\}$ from different viewpoints of an imaging sensor, where two keypoint sets $\{\mathcal{D}_p^1, \mathcal{D}_p^2 \}$ are extracted by local feature extraction method from $\mathcal{D}^1$ and $\mathcal{D}^2$ separately, the detector-based feature matching method $\mathcal{M}$ is supposed to estimate an assignment matrix $\mathbf{P}{\in}\left[ 0,1 \right] ^{M\times N}$ between $\mathcal{D}_p^1$ with $M$ keypoints and $\mathcal{D}_p^2$ with $N$ keypoints:
\begin{equation}
\mathbf{P} = \mathcal{M}(\mathcal{D}_p^1, \mathcal{D}_p^2).
\end{equation}

\noindent\textbf{Image Modality and Event Modality.}
Given a scene $\mathcal{L}(x,y,z,t)$, the RGB sensor records the integration of the brightness of the scene into a 2D tensor $I^{H\times W}$, whereas the event sensor records the polarity of brightness change for each pixel asynchronously and obtains an event stream $E{=}\{e_j\}{=}\{(x_j, y_j, t_j, p_j)\}$. Here, $(x_j, y_j)$ and $t_j$ represent the pixel coordinates and the timestamp of the event respectively, and $p_j{\in}\{-1, +1\}$ is the polarity.

\noindent\textbf{Inter-Modality Feature Matching.}
Due to the difference in the imaging principle, the data $E$ recorded through the event camera and the data $I$ through the RGB camera are with modality change, thus our work is built on the $\{E,I\}$ data.
Inter-modality feature matching addresses the challenges posed by changes in both viewpoint and modality that are observed during feature matching across these two sensor types. 
While traditional intra-modality feature matching methods primarily account for viewpoint changes, our model extends this to include modality changes as well.
The goal of inter-modality feature matching is as follows:
\begin{equation}
\mathbf{P}=\mathcal{M}(\mathcal{E}_E(E),\mathcal{E}_I(I)),
\label{eq:3}
\end{equation}
where $\mathcal{E}_E$ and $\mathcal{E}_I$ represent the local feature extractor for event data and image data respectively.

\subsection{Framework Overview}
\label{sec:framework overview}
To estimate the $\mathbf{P}$ in Eq.~(\ref{eq:3}), we split the whole estimation procedure into two separate stages, the local feature extraction stage and the feature matching stage.
As shown in Fig.~\ref{fig:framework}, in the local feature extraction stage, the data from two modalities are separately sent to the corresponding extractors.
For image modality, a well-learned $\mathcal{E}_I$ (such as SuperPoint~\cite{detone2018superpoint} and SiLK~\cite{gleize2023silk}) is adopted on the gray image $I^{H_I\times W_I\times 1}$, obtaining outputs $I_f^{H'\times W'\times C'}$ from backbone, $I_{score}^{H\times W\times 1}$ from score head and $I_{desc}^{H\times W\times C_d}$ from descriptor head.
For event modality, we choose events between $\{t_j{-}\Delta t,t_j\}$ where $t_j$ is the current timestamp and use a representation method (like voxelization) to construct an event tensor $E^{H_E\times W_E \times C_E}$.
Then the event tensor is sent to $\mathcal{E}_E$ and $E_f^{H'\times W'\times C'}$, $E_{score}^{H\times W\times 1}$ and $E_{desc}^{H\times W\times C_d}$ are obtained after the similar procedure as the image branch.

Noticing that the pixel where no event ever activated should not be treated as keypoint due to lack of information, an event mask $E_{mask}$ is accumulated from the event stream where the value is $0$ if no event ever happened in that pixel and $1$ vice versa.
The event score map $E_{score}$ is further multiplied with $E_{mask}$ before keypoint selection.

After the score maps and descriptor maps for both modalities are obtained, keypoint extraction is utilized to select sparse keypoints.
We follow the keypoint extraction procedure from SiLK~\cite{gleize2023silk}.
First, the points near the border are ignored.
Then the non-maximum suppression (NMS) is applied to sparsify the score map.
Finally, the positions $\mathbf{p}_i$ of the valid keypoints are extracted by performing a selection approach like thresholding or top-$k$ (which selects $k$ highest confidence features).
Once the $\mathbf{p}_i$ are detected, the corresponding descriptors $\mathbf{d}_i$ are queried from the descriptor map.
The same keypoints extraction is utilized for both modalities, resulting in two keypoint sets $E_p$ and $I_p$.

In the feature matching stage, the matcher $\mathcal{M}$ is supposed to estimate the $\mathbf{P}$.
$\mathcal{M}$ could be a simple Mutual Nearest Neighbor (MNN) which searches the nearest keypoint by calculating the L2 distance of the descriptors and utilizing a mutual check to have more precise inliers.
In addition, a network such as LightGlue~\cite{lindenberger2023lightglue} which utilizes Context Aggregation (CA) to aggregate descriptors from two keypoint sets through attention techniques could be a better choice if a training dataset with ground-truth depth and relative pose is available.

\subsection{Local Feature Distillation}
\label{sec:LFD}
The matcher $\mathcal{M}$ could work based on the assumption that the obtained keypoint sets $\{E_p, I_p\}$ are invariant to the observation $\mathcal{O}$, especially for no-biased matcher like MNN.
As mentioned in Sec.~\ref{sec:problem formulation}, the changes of $\mathcal{O}$ include both viewpoint change and modality change, and the feature space of $\mathcal{E}_E$ and $\mathcal{E}_I$ should be invariant to $\mathcal{O}$.

Noticing that there are many successful image extractors $\mathcal{E}_I$ showing great viewpoint invariance, one possible solution is that we force the event extractor $\mathcal{E}_E$ to have a similar feature space as existing $\mathcal{E}_I$.
In this case, $\mathcal{E}_E$ have the same viewpoint invariance property as $\mathcal{E}_I$, and since the $\mathcal{E}_E$ and $\mathcal{E}_I$ share a similar feature space, the modality invariance is achieved through the feature space alignment.

Based on this simple idea, we propose Local Feature Distillation (LFD) that transfers the knowledge from $\mathcal{E}_I$ to $\mathcal{E}_E$, by treating the feature space of $\mathcal{E}_I$ as the anchor and supervising $\mathcal{E}_I$ to be embedded within the same feature space.
In this case, the combined feature space of $\mathcal{E}_I$ and $\mathcal{E}_E$ is supposed to exhibit both viewpoint invariance and modality invariance.
To implement this, $I_{score}$ and $I_{desc}$ are selected for LFD, and $I_f$ is further used to restrict the latent space of the backbone.
In addition, as described in Sec.~\ref{sec:framework overview}, the pixels without event activated are not considered as keypoints.
As a result, we only supervise pixels that have event activated as indicated by $E_{mask}$.
Furthermore, to supervise the model at the pixel level, the pixel-level correspondence must be known.
This correspondence can be determined by a depth-aware system or a DAVIS camera that provides fully aligned events and images spatially.
As shown in Fig.~\ref{fig:framework}, three losses are adopted for supervision:
\begin{equation}
L_{feats}=\lVert E_f-I_f\rVert ^2_2,
\end{equation}
\begin{equation}
L_{score}=E_{mask}\lVert E_{score}-I_{score}\rVert ^2_2,
\end{equation}
\begin{equation}
L_{desc}=E_{mask}\lVert E_{desc}-I_{desc}\rVert _1.
\end{equation}
\noindent During training, the total local feature distillation loss $L_{lfd}$ combines the three losses above:
\begin{equation}
L_{lfd}=L_{feats}+L_{score}+L_{desc}.
\end{equation}

\subsection{Context aggregation and Matcher Training}
\label{sec: CA}
By utilizing LFD, the $\mathcal{E}_E$ and $\mathcal{E}_I$ are supposed to be aligned.
In this case, a simple MNN should work well.
However, the employment of LFD could not achieve such a unified feature space due to the modality difference.
In addition, the performance of $\mathcal{E}_E$ is constrained by the capabilities of $\mathcal{E}_I$ due to the distillation process.

To further improve the matching performance, a learnable network is considered to boost the feature matching performance.
During the local feature extraction stage, the inter-modality information is only learned through LFD losses, whereas context information of the data pair, which is important for feature matching, is not directly exchanged.
Observing that some Context Aggregation (CA) techniques~\cite{chen2021learning, lindenberger2023lightglue} which utilize GNN-based attention to propagate context information between a data pair have been successfully adopted for image feature matching, we propose to use a CA module like SGMNet~\cite{chen2021learning} or LightGlue~\cite{lindenberger2023lightglue} to transform the context information between event data and image data.
In addition, benefiting from the strong cross-modal information integration capability of attention mechanisms~\cite{xu2020cross, girdhar2023imagebind}, the inter-modality information could be better extracted, resulting in more satisfactory matching performance.

The learnable matcher $\mathcal{M}$ predicts every keypoint $(\mathbf{p}_i, \mathbf{d}_i)$ a reference descriptor $x_i$ and a matchability score $\sigma _i$ after CA:
\begin{equation}
    \{x_{i}^{E_p}, \sigma _i^{E_p}, x_{i}^{I_p}, \sigma _i^{I_p} \} = CA(E_p, I_p),
\end{equation}
\noindent then the similarity matrix $\mathbf{S}^{M\times N}$ is computed between $E_p$ and $I_p$:
\begin{equation}
    \mathbf{S}_{ij}=\left( x_{i}^{E_p} \right) ^{\top}x_{j}^{I_p}.
    \label{eq:9}
\end{equation}
\noindent After that, the assignment $\mathbf{P}$ could be obtained through dual-softmax~\cite{sun2021loftr}:
\begin{equation}
    \mathbf{P}_{ij}=\sigma _{i}^{E_p}\sigma _{j}^{I_p}\underset{k\in \left[ 0,|E_p| \right)}{\text{Soft}\max}\left( \mathbf{S}_{kj} \right) _i\underset{k\in \left[ 0,|I_p| \right)}{\text{Soft}\max}\left( \mathbf{S}_{ik} \right) _j.
\end{equation}
The CA module is trained under the supervision of the ground-truth depth map and relative pose.
Given the pixel-level depth map and relative pose, we project all the points from $E_p$ to the camera coordinates of $I_p$ and conversely.
The ground truth matches $\mathcal{M}$ are filtered based on a low reprojection error and a consistent depth.
Those unmatched points from $E_p$ and $I_p$ are marked as $\mathcal{A}$ and $\mathcal{B}$ respectively.
Then the negative likelihood loss $L_{nll}$ is utilized:
\begin{equation}
\begin{split}
L_{nll}=&-\frac{1}{|\mathcal{M}|}\sum_{\left( i,j \right) \in \mathcal{M}}{\log \mathbf{P}_{ij}} \\
&-\frac{1}{2|\mathcal{A}|}\sum_{i\in \mathcal{A}}{\log \left( 1-\sigma _{i}^{E_p} \right)} \\
&-\frac{1}{2|\mathcal{B}|}\sum_{j\in \mathcal{B}}{\log \left( 1-\sigma _{j}^{I_p} \right)}.
\end{split}
\end{equation}

\noindent When applying $L_{nll}$, the two local feature extractor $\mathcal{E}_E$ and $\mathcal{E}_I$ are frozen.

\section{Experiments}

\subsection{Datasets}

\noindent\textbf{MVSEC Dataset.}
MVSEC~\cite{zhu2018mvsec} records both event stream and grayscale images in both indoor and outdoor scenes.
The fully aligned events-image pair from DAVIS346 provides a dense correspondence map to enable the efficient training of the event extractor.
In addition, depth and accurate pose are also provided.
We use the \textit{indoor\_flying4} and \textit{outdoor\_day1} sequences to generate the MVSEC-RPE benchmark and the remaining sequences for training.

\noindent\textbf{EC Dataset.}
EC~\cite{mueggler2017event} provides several indoor scenes with simple patterns.
The DAVIS 240C camera is adopted for records and the ground-truth poses are also provided.
We use the split from DeepEvT~\cite{messikommer2023data} for training and testing.

\noindent\textbf{Generation of MVSEC-RPE and EC-RPE.}
Since there is no applicable benchmark for event-image matching, we propose the first relative pose estimation benchmark for event-image inter-modality local feature extraction and feature matching, building upon the MVSEC dataset and EC dataset.
To generate data pairs from the sequences of MVSEC, we perform the same sampling rules as \cite{sarlin2020superglue}, which calculates the overlap score between two images with different timestamps, and filters out those pairs with overlap score not in $[0.4, 0.8]$.
Additionally, the ground truth matches of the data pair are used for relative pose estimation with RANSAC~\cite{fischler1981random}. 
Only the data pairs that achieve an angular error of less than $1$ degree for both $R$ and $t$, and an inlier ratio higher than $0.9$ are selected as reliable test pairs.
Based on the sampling rules above, $600$ data pairs in total are randomly generated.
For EC, there is no available depth provided, so we randomly sample $198$ non-repeating test pairs from test sequences.

\subsection{Implementation Details}

\noindent\textbf{Model Setup.}
In the local feature extraction stage, two types of image extractor $\mathcal{E}_I$, \ie, SuperPoint~\cite{detone2018superpoint} and SiLK~\cite{gleize2023silk}, are selected for training the VGG-based event extractor $\mathcal{E}_E$.
The training data pair $\{E, I\}$ is randomly selected with the same timestamp, providing fully aligned points from different modalities for efficient training.
Furthermore, the events and images are rectified, as the SuperPoint and SiLK are trained on non-distortion image data.

In the feature matching stage, two matching methods, MNN and LightGlue (LG)~\cite{lindenberger2023lightglue} are tested.
Since the official LG is trained upon the keypoints of SuperPoint, we compare the performance of the fine-tuned model and the model trained from scratch.
For SiLK, no trained LG is valid, so the training is from scratch.
During training, the event data $E$ is selected only when the current timestamp $t_i$ has corresponding image and depth data.
One image $I$ is then indexed with an index difference in $[0, 60]$ compared with the index of the corresponding image of $E$.

For both training stages of MVSEC, the $\Delta t$ of the event stream is set to $0.4s$, and the event stream in $[t_i{-}\Delta t, t_i]$ is transformed to a voxel with $16$ channels.
During keypoint extraction, we remove the image border of $4$, use an NMS radius of $4$, and then utilize the top-$1024$ criteria to select $1024$ keypoints with the highest score.
For EC, the $\Delta t$ is set to $0.04s$ to have a similar event rate per pixel, and we utilize top-$512$ due to the low resolution of DAVIS 240C.

\noindent\textbf{Baselines.}
We primarily compare our method with two intra-modality approaches, which employ event-to-video methods (E2VID+~\cite{stoffregen2020reducing} and HyperE2VID~\cite{ercan2024hypere2vid}) through the EVREAL toolkit~\cite{ercan2023evreal} followed by intra-modality local feature extraction and matching.
The event stream of each sequence is transformed into an image sequence, and then the image-oriented local feature extraction and matching are performed on the transformed image and the original image.
Furthermore, to compare with real intra-modality feature matching, image-image feature matching is performed as the reference, which directly uses the two images provided by the DAVIS camera for feature matching.

\begin{figure*}[t!]
    \centering
    \includegraphics[width=\linewidth]{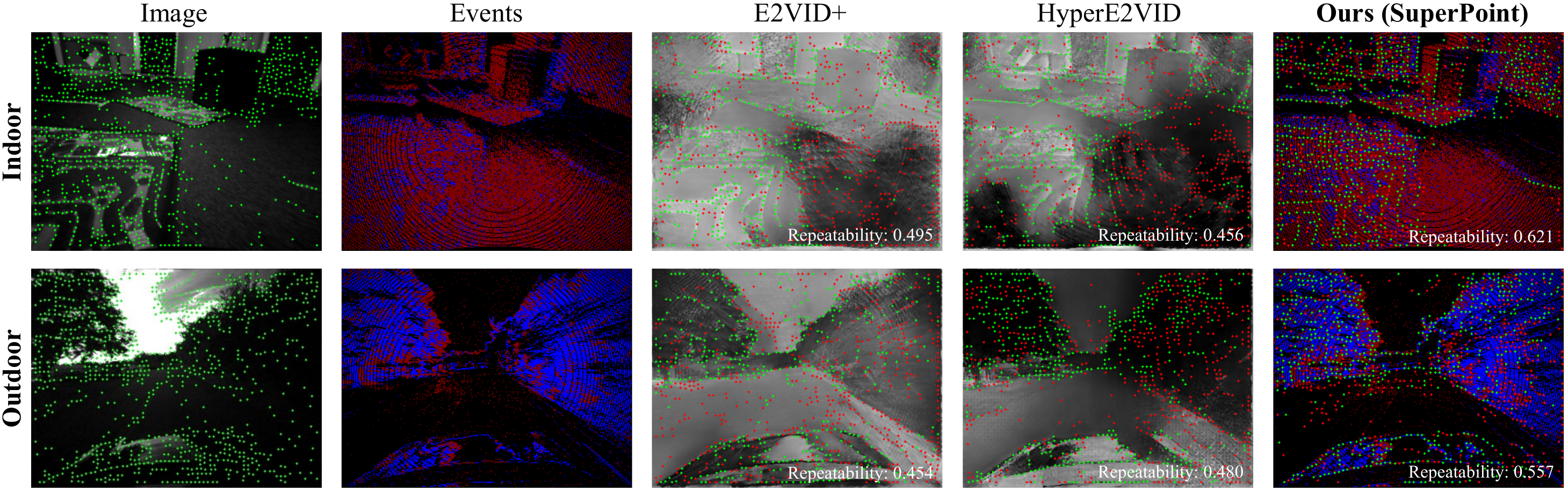}
    \caption{\textbf{Qualitative results of keypoint similarity.} Keypoints that satisfy the \emph{Repeatability} criterion with $\epsilon{=}3$ are shown in \textcolor{green}{green} in the three rightmost columns, while the rest are in \textcolor{red}{red}. The \emph{Repeatability} score for each method is marked at the bottom of the image. The event-to-video methods suffer from artifacts or inconsistent dynamic range, resulting in low \emph{Repeatability}.}
    \label{fig:keypoint_similarity}
\end{figure*}

\subsection{Cross-modal Keypoint Similarity Evaluation}

\noindent\textbf{Setup:} We use the fully aligned events and image from MVSEC-RPE to extract keypoints separately, and then calculate \textit{Repeatability}, valid descriptor distance (\textit{VDD}), and valid descriptor angle (\textit{VDA}).
In this case, the evaluation solely focuses on assessing the cross-modal consistency of the keypoints.
When points from the two points set have a spatial distance under $\epsilon$, they are treated as valid prediction pairs.
\textit{Repeatability} measures the ratio of valid predictions to the total number of predictions. 
Meanwhile, \textit{VDD} and \textit{VDA} calculate the L2 distance and the vector angle between descriptors from each prediction pair respectively.

\begin{table}[t]
\large
\centering
\renewcommand{\arraystretch}{1.3}
\resizebox{1.0\columnwidth}{!}{
\setlength{\tabcolsep}{3pt}

\begin{tabular}{l|lccccccc}
\hline
\multirow{2}{*}{\begin{tabular}[c]{@{}c@{}}\textbf{Extractor}\end{tabular}} & \multicolumn{1}{l}{\multirow{2}{*}{\textbf{Method}}} & \multirow{2}{*}{\textbf{Modality}} & \multicolumn{2}{c}{\underline{\textbf{Repeatability} $\uparrow$}} & \multicolumn{2}{c}{\underline{\hspace{0.5cm}\textbf{VDD} $\downarrow$\hspace{0.5cm}}} & \multicolumn{2}{c}{\underline{\hspace{0.5cm}\textbf{VDA} $\downarrow$\hspace{0.5cm}}} \\ 
 & \multicolumn{1}{c}{} &  & $\epsilon = 1$ & $\epsilon = 3$ & $\epsilon = 1$ & $\epsilon = 3$ & $\epsilon = 1$ & $\epsilon = 3$ \\ \hline \hline
\multirow{3}{*}{SuperPoint~\cite{detone2018superpoint}} & E2VID+~\cite{stoffregen2020reducing} & \textit{Intra.} & 0.141 & 0.515 & 0.924 & 0.953 & 55.725 & 57.605 \\
 & HyperE2VID~\cite{ercan2024hypere2vid} & \textit{Intra.}  & 0.145 & 0.517 & 0.932 & 0.954 & 56.229 & 57.611 \\
 & \cellcolor{gray!20}Ours (SuperPoint) & \cellcolor{gray!20}\textit{Inter.} & \cellcolor{gray!20}\textbf{0.178} & \cellcolor{gray!20}\textbf{0.528} & \cellcolor{gray!20}\textbf{0.724} & \cellcolor{gray!20}\textbf{0.770} & \cellcolor{gray!20}\textbf{42.938} & \cellcolor{gray!20}\textbf{45.843} \\ \hline 
\multirow{3}{*}{SiLK~\cite{gleize2023silk}} & E2VID+~\cite{stoffregen2020reducing} & \textit{Intra.} & 0.107 & 0.469 & 1.597 & 1.698 & 69.542 & 74.537 \\
 & HyperE2VID~\cite{ercan2024hypere2vid} & \textit{Intra.}  & 0.098 & 0.428 & 1.617 & 1.709 & 70.544 & 75.137 \\ 
 & \cellcolor{gray!20}Ours (SiLK) & \cellcolor{gray!20}\textit{Inter.} & \cellcolor{gray!20}\textbf{0.156} & \cellcolor{gray!20}\textbf{0.557} & \cellcolor{gray!20}\textbf{1.466} & \cellcolor{gray!20}\textbf{1.630} & \cellcolor{gray!20}\textbf{63.299} & \cellcolor{gray!20}\textbf{71.196} \\ \hline
\end{tabular}

}
\caption{\textbf{Keypoint similarity} between two modalities on \textit{MVSEC-RPE} test set with only modality change. \textit{Intra} represents \textit{intra-modality} and \textit{Inter} represents \textit{inter-modality}.}
\label{table:keypoints similarity}
\vskip -1.\baselineskip plus -1fil
\end{table}

\noindent\textbf{Results:} As shown in Table~\ref{table:keypoints similarity}, our framework exhibits much higher \textit{Repeatability}, lower \textit{VDD} and \textit{VDA} than the intra-modality pipelines with both image extractors.
It is also noteworthy that the event-to-video methods follow an iterative manner to use the information of the whole sequence, whereas our framework only uses an events slice but outperforms them.
As illustrated in Fig.~\ref{fig:keypoint_similarity}, due to the presence of artifacts or inconsistencies in dynamic range, the explicit modality transformation methods are unable to achieve high levels of \emph{Repeatability}, whereas our proposed framework maintains its superiority.

\begin{table*}[!t]
\centering
\renewcommand{\arraystretch}{1.1}
\setlength{\tabcolsep}{10pt}
\resizebox{0.95\linewidth}{!}{

\begin{tabular}{l|l|llccccccc}
\hline
\multirow{2}{*}{\begin{tabular}[c]{@{}c@{}} \textbf{Extractor}\end{tabular}} & \multirow{2}{*}{\textbf{Modality}} & \multicolumn{1}{l}{\multirow{2}{*}{\textbf{Method}}} & \multicolumn{1}{c}{\multirow{2}{*}{\textbf{Matcher}}} & \multirow{2}{*}{\textbf{RPE Inlier} $\uparrow$} & \multicolumn{3}{c}{\underline{\hspace{1cm}\textbf{RPE Ratio} $\uparrow$\hspace{1cm}}} & \multicolumn{3}{c}{\underline{\hspace{1cm}\textbf{RPE AUC} $\uparrow$\hspace{1cm}}} \\
 &  & \multicolumn{1}{c}{} & \multicolumn{1}{c}{} &  & $\epsilon=5\degree$ & $\epsilon=10\degree$ & $\epsilon=20\degree$ & $\epsilon=5\degree$ & $\epsilon=10\degree$ & $\epsilon=20\degree$ \\ \hline \hline
\multirow{12}{*}{SuperPoint~\cite{detone2018superpoint}} & \multirow{9}{*}{Intra-Modality} & \multirow{3}{*}{\textcolor{gray!80}{Image-Image}} & \textcolor{gray!80}{MNN} & \textcolor{gray!80}{0.252} & \textcolor{gray!80}{0.133} & \textcolor{gray!80}{0.276} & \textcolor{gray!80}{0.333} & \textcolor{gray!80}{4.30} & \textcolor{gray!80}{12.67} & \textcolor{gray!80}{22.04} \\
 &  &  & \textcolor{gray!80}{LG (f.s.)} & \textcolor{gray!80}{0.415} & \textcolor{gray!80}{0.328} & \textcolor{gray!80}{0.675} & \textcolor{gray!80}{0.856} & \textcolor{gray!80}{11.35} & \textcolor{gray!80}{32.27} & \textcolor{gray!80}{55.97} \\
 &  &  & \cellcolor{gray!20}\textcolor{gray!80}{LG (f.t.)} & \cellcolor{gray!20}\textcolor{gray!80}{0.476} & \cellcolor{gray!20}\textcolor{gray!80}{0.463} & \cellcolor{gray!20}\textcolor{gray!80}{0.870} & \cellcolor{gray!20}\textcolor{gray!80}{0.948} & \cellcolor{gray!20}\textcolor{gray!80}{17.80} & \cellcolor{gray!20}\textcolor{gray!80}{44.81} & \cellcolor{gray!20}\textcolor{gray!80}{68.85} \\ \cline{3-11} 
 &  & \multirow{3}{*}{E2VID+~\cite{stoffregen2020reducing}} & MNN & 0.155 & 0.088 & 0.245 & 0.371 & 2.95 & 10.45 & 20.95 \\
 &  &  & LG (f.s.) & 0.404 & 0.280 &0.603 & 0.821 & 9.40 & 28.14 & 50.90 \\
 &  &  & \cellcolor{gray!20}LG (f.t.) & \cellcolor{gray!20}0.454 & \cellcolor{gray!20}0.216 & \cellcolor{gray!20}0.535 & \cellcolor{gray!20}0.818 & \cellcolor{gray!20}8.03 & \cellcolor{gray!20}23.18 & \cellcolor{gray!20}47.33 \\ \cline{3-11} 
 &  & \multirow{3}{*}{HyperE2VID~\cite{ercan2024hypere2vid}} & MNN & 0.145 & 0.091 & 0.218 & 0.348 & 3.24 & 9.31 & 19.38 \\
 &  &  & LG (f.s.) & 0.385 & 0.168 & 0.481 & 0.756 & 5.86 & 19.64 & 41.71 \\
 &  &  & \cellcolor{gray!20}LG (f.t.) & \cellcolor{gray!20}0.457 & \cellcolor{gray!20}0.201 & \cellcolor{gray!20}0.523 & \cellcolor{gray!20}0.776 & \cellcolor{gray!20}6.70 & \cellcolor{gray!20}21.43 & \cellcolor{gray!20}44.41 \\ \cline{2-11} 
 & \multirow{3}{*}{Inter-Modality} & \multirow{3}{*}{\textbf{Ours (SuperPoint)}} & MNN & 0.227 & 0.115 & 0.276 & 0.401 & 3.75 & 12.14 & 24.00 \\
 &  &  & LG (f.s.) & \underline{0.497} & \underline{0.403} & \underline{0.763} & \underline{0.898} & \underline{14.55} & \underline{38.03} & \underline{61.72} \\
 &  &  & \cellcolor{gray!20}LG (f.t.) & \cellcolor{gray!20}\textbf{0.574} & \cellcolor{gray!20}\textbf{0.585} & \cellcolor{gray!20}\textbf{0.868} & \cellcolor{gray!20}\textbf{0.931} & \cellcolor{gray!20}\textbf{24.76} & \cellcolor{gray!20}\textbf{50.42} & \cellcolor{gray!20}\textbf{70.91} \\ \hline 
\multirow{8}{*}{SiLK~\cite{gleize2023silk}} & \multirow{6}{*}{Intra-Modality} & \multirow{2}{*}{\textcolor{gray!80}{Image-Image}} & \textcolor{gray!80}{MNN} & \textcolor{gray!80}{0.207} & \textcolor{gray!80}{0.121} & \textcolor{gray!80}{0.281} & \textcolor{gray!80}{0.341} & \textcolor{gray!80}{3.70} & \textcolor{gray!80}{12.28} & \textcolor{gray!80}{22.03} \\
 &  &  & \cellcolor{gray!20}\textcolor{gray!80}{LG (f.s.)} & \cellcolor{gray!20}\textcolor{gray!80}{0.426} & \cellcolor{gray!20}\textcolor{gray!80}{0.300} & \cellcolor{gray!20}\textcolor{gray!80}{0.635} & \cellcolor{gray!20}\textcolor{gray!80}{0.800} & \cellcolor{gray!20}\textcolor{gray!80}{10.12} & \cellcolor{gray!20}\textcolor{gray!80}{29.53} & \cellcolor{gray!20}\textcolor{gray!80}{52.03} \\ \cline{3-11} 
 &  & \multirow{2}{*}{E2VID+~\cite{stoffregen2020reducing}} & MNN & 0.046 & 0.038 & 0.105 & 0.281 & 1.22 & 4.21 & 11.98 \\
 &  &  & \cellcolor{gray!20}LG (f.s.) & \cellcolor{gray!20}\underline{0.365} & \cellcolor{gray!20}0.101 & \cellcolor{gray!20}0.321 & \cellcolor{gray!20}0.646 & \cellcolor{gray!20}3.04 & \cellcolor{gray!20}12.09 & \cellcolor{gray!20}31.64 \\ \cline{3-11} 
 &  & \multirow{2}{*}{HyperE2VID~\cite{ercan2024hypere2vid}} & MNN & 0.044 & 0.046 & 0.101 & 0.260 & 1.16 & 4.33 & 11.27 \\
 &  &  & \cellcolor{gray!20}LG (f.s.) & \cellcolor{gray!20}0.353 & \cellcolor{gray!20}\underline{0.135} & \cellcolor{gray!20}\underline{0.421} & \cellcolor{gray!20}\underline{0.720} & \cellcolor{gray!20}\underline{3.76} & \cellcolor{gray!20}\underline{16.05} & \cellcolor{gray!20}\underline{38.11} \\ \cline{2-11} 
 & \multirow{2}{*}{Inter-Modality} & \multirow{2}{*}{\textbf{Ours (SiLK)}} & MNN & 0.113 & 0.090 & 0.226 & 0.411 & 2.93 & 9.74 & 21.32 \\
 &  &  & \cellcolor{gray!20}LG (f.s.) & \cellcolor{gray!20}\textbf{0.492} & \cellcolor{gray!20}\textbf{0.351} & \cellcolor{gray!20}\textbf{0.726} & \cellcolor{gray!20}\textbf{0.891} & \cellcolor{gray!20}\textbf{12.27} & \cellcolor{gray!20}\textbf{34.98} & \cellcolor{gray!20}\textbf{59.05} \\ \hline
\end{tabular}
}
\caption{\textbf{Relative pose estimation results on MVSEC-RPE benchmark.} (f.s.) represents the model trained from scratch, and (f.t.) represents the fine-tuned model.}
\label{table:rpe}
\vskip -0.5\baselineskip plus -1fil
\end{table*}

\begin{figure*}[!t]
    \centering
    \includegraphics[width=\linewidth]{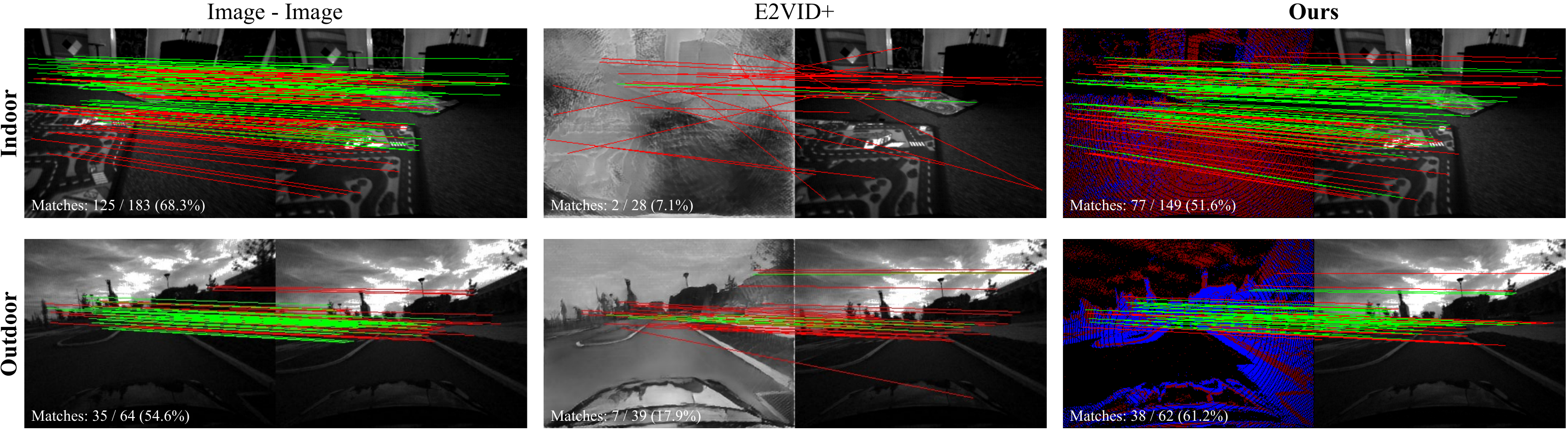}
    \caption{\textbf{Qualitative results of matching results on the MVSEC-RPE dataset.} SuperPoint is employed as the image extractor, while the fine-tuned LightGlue is utilized to estimate the matches. Correct matches are indicated by \textcolor{green}{green lines} and mismatches are by \textcolor{red}{red lines}.}
    \label{fig:matching results}
    \vskip -1.0\baselineskip plus -1fil
\end{figure*}

\begin{table}[t]
\large
\centering
\renewcommand{\arraystretch}{1.1}
\resizebox{1.0\columnwidth}{!}{
\setlength{\tabcolsep}{6pt}

\begin{tabular}{l|l|c|cc}
\hline
\multicolumn{1}{c|}{\multirow{2}{*}{\begin{tabular}[c]{@{}c@{}}\textbf{Extractor}\end{tabular}}} & \multicolumn{1}{l|}{\multirow{2}{*}{\textbf{Method}}} & \multirow{2}{*}{\begin{tabular}[c]{@{}c@{}}\textbf{RPE}\\ \textbf{Inlier} $\uparrow$\end{tabular}} & \textbf{RPE Ratio} $\uparrow$ & \textbf{RPE AUC} $\uparrow$ \\ \cline{4-5} 
\multicolumn{1}{c|}{} & \multicolumn{1}{c|}{} &  & $\epsilon = 5\degree/10\degree/20\degree$ & $\epsilon = 5\degree/10\degree/20\degree$ \\ \hline \hline
\multirow{4}{*}{SuperPoint} & \textcolor{gray!80}{Image-Image} & \textcolor{gray!80}{0.331} & \textcolor{gray!80}{0.055/0.151/0.287} & \textcolor{gray!80}{2.31/6.81/15.56} \\
 & E2VID+ & \underline{0.232} & \underline{0.030}/\underline{0.095}/\underline{0.156} & \textbf{1.79}/\underline{4.63}/\underline{8.89} \\
 & HyperE2VID & 0.219 & 0.015/0.060/0.136 & 0.99/2.51/6.53 \\ 
 & \cellcolor{gray!20}\textbf{Ours (SuperPoint)} & \cellcolor{gray!20}\textbf{0.285} & \cellcolor{gray!20}\textbf{0.040}/\textbf{0.121}/\textbf{0.237} & \cellcolor{gray!20}\underline{1.49}/\textbf{5.12}/\textbf{12.21} \\ \hline 
\multirow{4}{*}{SiLK} & \textcolor{gray!80}{Image-Image} & \textcolor{gray!80}{0.276} & \textcolor{gray!80}{0.070/0.161/0.292} & \textcolor{gray!80}{2.64/7.47/15.63} \\
 & E2VID+ & \underline{0.111} & \underline{0.015}/\underline{0.045}/0.090 & \underline{0.86}/\textbf{1.92}/4.57 \\
 & HyperE2VID & 0.092 & 0.005/\textbf{0.050}/\underline{0.095} & 0.32/1.85/\underline{4.80} \\ 
 & \cellcolor{gray!20}\textbf{Ours (SiLK)} & \cellcolor{gray!20}\textbf{0.190} & \cellcolor{gray!20}\textbf{0.020}/0.030/\textbf{0.111} & \cellcolor{gray!20}\textbf{0.97}/\underline{1.71}/\textbf{4.88} \\ \hline
\end{tabular}
}
\vskip -0.3\baselineskip plus -1fil
\caption{\textbf{Relative pose estimation results on the EC-RPE benchmark.} MNN is employed during feature matching.}
\label{table:ec rpe}
\vskip -0.5\baselineskip plus -1fil
\end{table}

\subsection{Relative Pose Estimation Results}

\noindent\textbf{Setup:} When conducting relative pose estimation, we use the generated data pairs from MVSEC-RPE and EC-RPE, where each pair contains an event stream and an image in different timestamps.
We use the matched points predicted from the model to estimate an essential matrix with the vanilla RANSAC~\cite{fischler1981random}, and then decompose it into a rotation and a translation.
The RANSAC is applied for all tested models, and the \emph{Inlier Ratio} of RANSAC is used to evaluate the robustness of the predicted matching results.

Pose error is computed as the maximum angular error in rotation and translation.
Within a specified threshold $\epsilon$, the \emph{RPE Ratio} reports the ratio of posture errors below $\epsilon$. 
Additionally, the \emph{RPE AUC} reports the area under curve (AUC) of the pose error below $\epsilon$.

\noindent\textbf{Results:} As shown in Table~\ref{table:rpe} and Table~\ref{table:ec rpe}, our framework surpasses all explicit transformation methods with substantial superiority in all compared conditions.
Results obtained by performing CA are consistently better than applying MNN.
Compared with the image-image feature matching approach that has no modality difference, our framework achieves a marginally worse performance when applying MNN but outperforms a lot when the CA is utilized.
This demonstrates that LFD provides the inter-modality prior for local feature extraction, and the usage of CA results in a preferable performance building on that prior.
The matching results in MVSEC-RPE are illustrated in Fig.~\ref{fig:matching results}.
Our framework provides more consistent feature matches compared to E2VID+ and achieves comparable or even superior performance to unmediated image-to-image matching.

We notice that E2VID+ yields better RPE results when the LG is trained from scratch rather than fine-tuned from the official pre-trained model.
This shows the images generated by event-to-video methods are not necessarily suitable for fine-tuning image tasks, as the difference in probability distribution between these images and real images is difficult to measure.
Our framework, instead, learns the latent feature distribution from an image extractor trained on real images, thereby incorporating task-relevant properties of real images, which is the reason why improvements are observed in fine-tuning LG.

\subsection{Ablation Studies}

\begin{figure}[t]
    \resizebox{0.48\linewidth}{!}{
        \begin{tikzpicture}
            \begin{axis}[
                xtick={50, 100, 200, 400, 800}, %
                % legend pos=south west,
                xticklabels={50, 100, 200, 400, 800}, %
                ymin=0.135,
                grid=both,
                xmode=log,
                grid style={line width=.1pt, draw=gray!10},
                major grid style={line width=.2pt,draw=gray!50},
                minor tick num=2,
                % axis x line*=bottom,
                % axis y line*=left,
                height=1.0\linewidth,
                width=1.0\linewidth,
                ylabel style= {align=center, font=\large},
                xlabel style = {font=\large, font=\large},
                ylabel={Repeatability $\uparrow$ ($\epsilon{=}1$)},
                xlabel={$\Delta t$ during inference (ms)},
                yticklabel style = {font=\large},
                xticklabel style = {font=\large},
                legend to name={mylegend},
                legend columns=-1,
                legend entries={50ms, 100ms, 200ms, 400ms, 800ms, HyperE2VID, E2VID+},
                legend style={/tikz/every even column/.append style={column sep=0.2cm}, at={(1.0, 1.15)}, text=black}
            ]
            % 100ms
            \addplot[mark=o, very thick, gray!60, mark options={solid}, line width=1pt, mark size=3pt] plot coordinates {
                (50, 0.158) %
                (100, 0.168) %
                (200, 0.166) %
                (400, 0.156) %
                (800, 0.142) %
            };
            % \addlegendentry{100ms}
            % 200ms
            \addplot[mark=square, very thick, gray!60, mark options={solid}, line width=1pt, mark size=3pt %
                    ] plot coordinates {
                (50, 0.154) %
                (100, 0.166) %
                (200, 0.174) %
                (400, 0.170) %
                (800, 0.160) %
            };
            % \addlegendentry{200ms}
            % 400ms
            \addplot[mark=triangle, very thick, red, mark options={solid}, line width=1pt, mark size=3pt %
                    ] plot coordinates {
                (50, 0.153) %
                (100, 0.163) %
                (200, 0.173) %
                (400, 0.178) %
                (800, 0.172) %
            };
            % \addlegendentry{400ms}
            % 800ms
            \addplot[mark=star, very thick, gray!60, mark options={solid}, line width=1pt, mark size=3pt %
                    ] plot coordinates {
                (50, 0.147) %
                (100, 0.161) %
                (200, 0.171) %
                (400, 0.178) %
                (800, 0.179) %
            };
            % \addlegendentry{800ms}
            % HyperE2VID
            \addplot[mark=otimes, very thick, brown, mark options={solid}, line width=1pt, mark size=3pt %
                    ] plot coordinates {
                (50, 0.145) %
                (100, 0.145) %
                (200, 0.145) %
                (400, 0.145) %
                (800, 0.145) %
            };
            % \addlegendentry{HyperE2VID}
            % E2VID+
            \addplot[mark=otimes, very thick, cyan, mark options={solid}, line width=1pt, mark size=3pt %
                    ] plot coordinates {
                (50, 0.141) %
                (100, 0.141) %
                (200, 0.141) %
                (400, 0.141) %
                (800, 0.141) %
            };
            % \addlegendentry{E2VID+}
            % \legend{100ms, 200ms, 400ms, 800ms, HyperE2VID, E2VID+}
        
            \end{axis}
    
        \end{tikzpicture}
    }
    % \hfill
    \resizebox{0.48\linewidth}{!}{
        \begin{tikzpicture}
            \begin{axis}[
                xtick={50, 100, 200, 400, 800}, %
                % legend pos=south west,
                xticklabels={50, 100, 200, 400, 800}, %
                ymin=40,
                grid=both,
                xmode=log,
                grid style={line width=.1pt, draw=gray!10},
                major grid style={line width=.2pt,draw=gray!50},
                minor tick num=2,
                % axis x line*=bottom,
                % axis y line*=left,
                height=0.96\linewidth,
                width=1.0\linewidth,
                ylabel style= {align=center, font=\large},
                xlabel style = {font=\large, font=\large},
                ylabel={VDA $\downarrow$ ($\epsilon{=}1$)},
                xlabel={$\Delta t$ during inference (ms)},
                yticklabel style = {font=\large},
                xticklabel style = {font=\large},
            ]
            % 100ms
            \addplot[mark=o, very thick, gray!60, mark options={solid}, line width=1pt, mark size=3pt] plot coordinates {
                (50, 46.9237) %
                (100, 44.1387) %
                (200, 45.8651) %
                (400, 48.1301) %
                (800, 51.8376) %
            };
            % \addlegendentry{100ms}
            % 200ms
            \addplot[mark=square, very thick, gray!60, mark options={solid}, line width=1pt, mark size=3pt %
                    ] plot coordinates {
                (50, 48.6874) %
                (100, 45.6423) %
                (200, 43.2460) %
                (400, 45.1350) %
                (800, 47.2586) %
            };
            % % \addlegendentry{200ms}
            % 400ms
            \addplot[mark=triangle, very thick, red, mark options={solid}, line width=1pt, mark size=3pt %
                    ] plot coordinates {
                (50, 50.5271) %
                (100, 47.3179) %
                (200, 45.0074) %
                (400, 42.938) %
                (800, 44.2931) %
            };
            % \addlegendentry{400ms}
            % 800ms
            \addplot[mark=star, very thick, gray!60, mark options={solid}, line width=1pt, mark size=3pt %
                    ] plot coordinates {
                (50, 53.2844) %
                (100, 49.1056) %
                (200, 46.3753) %
                (400, 44.2653) %
                (800, 43.4455) %
            };
            % \addlegendentry{800ms}
            % HyperE2VID
            \addplot[mark=otimes, very thick, brown, mark options={solid}, line width=1pt, mark size=3pt %
                    ] plot coordinates {
                (50, 56.229) %
                (100, 56.229) %
                (200, 56.229) %
                (400, 56.229) %
                (800, 56.229) %
            };
            % \addlegendentry{HyperE2VID}
            % E2VID+
            \addplot[mark=otimes, very thick, cyan, mark options={solid}, line width=1pt, mark size=3pt %
                    ] plot coordinates {
                (50, 55.725) %
                (100, 55.725) %
                (200, 55.725) %
                (400, 55.725) %
                (800, 55.725) %
            };
            % \addlegendentry{E2VID+}
        
            \end{axis}
    
        \end{tikzpicture}
    }
    \resizebox{1.0\linewidth}{!}{
    \ref{mylegend}}
    \vskip -0.5\baselineskip plus -1fil
    \caption{\textbf{Temporal invariance test.} We assess keypoint similarity under different $\Delta t$ values during inference. Each line represents a model trained with a $\Delta t$. Explicit modality transformation methods are also included. Since the entire sequence of events is used for the modal transformation, the results of event-to-video methods are identical across different values of $\Delta t$.}
    \label{fig:temporal_property}
    % \vskip -1.5\baselineskip plus -1fil
\end{figure}
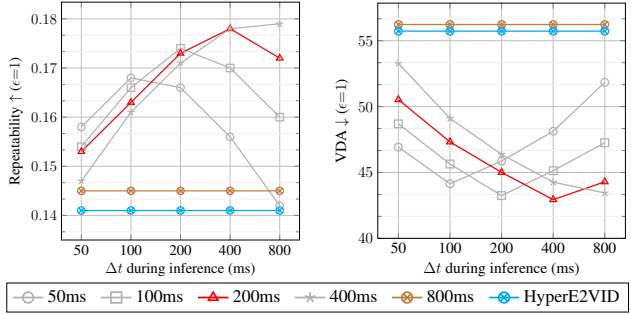

\noindent\textbf{Temporal Invariance.}
As shown in Fig.~\ref{fig:temporal_property}, we set four ${\Delta}t$ choices for training and five ${\Delta}t$ during inference.
The \emph{Repeatability} and \emph{VDA} results with $\epsilon{=}1$ are used for comparison. 
The results show that the best performance always appears when ${\Delta}t$ in training and inference are the same, and as the training ${\Delta}t$ increases, the best performance turns out an upward trend.
When ${\Delta}t$ comes to $400ms$ and higher, the best performance could not be better, which shows more information can lead to better results, but there are inherent limitations to the extent of performance improvements that can be achieved.
In addition, although the performance is not that great when ${\Delta}t$ in training and inference are not the same, it is significantly better than the explicit transformation methods in most instances.

% \begin{table}[t]
% \Large
% \centering
% \renewcommand{\arraystretch}{1.1}
% \tabcolsep=10pt
% \resizebox{0.7\linewidth}{!}{
% % \setlength{\tabcolsep}{10pt}

% \begin{tabular}{l|ccc}
% \hline
% \multicolumn{1}{c|}{Representaion} & Repeatability $\uparrow$ & VDD $\downarrow$ & VDA $\downarrow$ \\ \hline \hline
% E2VID+ & 0.141 & 0.924 & 55.725 \\
% HyperE2VID & 0.145 & 0.932 & 56.229 \\ \hline \hline
% Event Stack & 0.155 & 0.737 & 43.804 \\
% Time Surface & 0.150 & 0.762 & 45.302 \\
% Voxel Grid & \textbf{0.178} & \textbf{0.724} & \textbf{42.938} \\ \hline
% \end{tabular}
% }
% \caption{\textbf{Results of utilizing different event representation choices.} The three representation choices all surpass the event-to-video approaches on keypoint similarity.}
% \label{table:event_reprentation}
% \vskip -1.2\baselineskip plus -1fil
% \end{table}

\begin{table}[t]
\Large
\centering
\renewcommand{\arraystretch}{1.1}
\tabcolsep=10pt
\resizebox{0.7\linewidth}{!}{

\begin{tabular}{l|ccc}
\hline
\multicolumn{1}{c|}{\textbf{Representaion}} & \textbf{Repeatability} $\uparrow$ & \textbf{VDD} $\downarrow$ & \textbf{VDA} $\downarrow$ \\ \hline \hline
E2VID+ & 0.141 & 0.924 & 55.725 \\
HyperE2VID & 0.145 & 0.932 & 56.229 \\ \hline
Event Stack & 0.155 & 0.737 & 43.804 \\
Time Surface & 0.150 & 0.762 & 45.302 \\
\cellcolor{gray!20}Voxel Grid & \cellcolor{gray!20}\textbf{0.178} & \cellcolor{gray!20}\textbf{0.724} & \cellcolor{gray!20}\textbf{42.938} \\ \hline
\end{tabular}
}
\caption{\textbf{Results of utilizing different event representation choices.} The three representation choices all surpass the event-to-video approaches on keypoint similarity.}
\label{table:event_reprentation}
% \vskip -0.8\baselineskip plus -1fil
\end{table}

\noindent\textbf{Event Representation Choices.}
We test three event representation types in EI-Nexus, including Event Stack~\cite{wang2019event}, Time Surface~\cite{mueggler2017fast}, and Voxel Grid~\cite{zhu2019unsupervised}.
As shown in Table~\ref{table:event_reprentation}, the Voxel Grid performs the best in all three metrics, and both Event Stack and Time Surface outperform the explicit transformation method by a large margin.
It demonstrates that any reasonable event representation could benefit from LFD and the learned event extractor consistently achieves better performance.
Furthermore, the performance of models using different event representations varies, suggesting that appropriate representation selection is crucial for effective inter-modality local feature extraction.

\begin{table}[t]
\Large
\centering
\renewcommand{\arraystretch}{1.1}
\resizebox{0.7\linewidth}{!}{

\begin{tabular}{ccc|ccc}
\hline
$L_{feats}$ & $L_{score}$ & $L_{desc}$ & \textbf{Repeatability} $\uparrow$ & \textbf{VDD} $\downarrow$ & \textbf{VDA} $\downarrow$ \\ \hline \hline
 & \checkmark &  & \textbf{0.180} & 1.410 & 89.72 \\
 &  & \checkmark & 0.073 & 0.766 & 45.69 \\
 & \checkmark & \checkmark & 0.134 & \underline{0.754} & \underline{44.93} \\
% \checkmark & \checkmark &  & \textbf{0.180} & 1.412 & 89.89 \\
% \checkmark &  & \checkmark & 0.075 & 0.766 & 45.61 \\
\cellcolor{gray!20}\checkmark & \cellcolor{gray!20}\checkmark & \cellcolor{gray!20}\checkmark & \cellcolor{gray!20}\underline{0.178} & \cellcolor{gray!20}\textbf{0.724} & \cellcolor{gray!20}\textbf{42.94} \\ \hline
\end{tabular}
}
\caption{\textbf{Ablation studies for $L_{lfd}$}.}
\label{table:losses_ablation}
\vskip -0.8\baselineskip plus -1fil
\end{table}
\noindent\textbf{Components in $L_{lfd}$.}
The LFD process for the detector head and the descriptor head is considered a multi-task learning problem.
We test all reasonable combinations of $L_{lfd}$ to fully understand its behavior.
The results shown in Table~\ref{table:losses_ablation} demonstrate that the optimization of $L_{score}$ and $L_{desc}$ is balanced.
However, with the participation of $L_{feats}$, the joint latent space of the two heads is constrained, thus providing more satisfying keypoint extraction results.

% \section{Discussion}
% \input{Tex_content/discussion}

\section{Conclusion}
In this paper, we present a detector-based inter-modality local feature extraction and matching framework for event-image data.
The proposed framework employs local feature distillation to transfer the local feature knowledge from a well-learned image extractor to the event extractor.
The framework also supports various feature-matching methods, including MNN and LightGlue.
To provide a benchmark for event-image feature matching, we construct the first benchmark MVSEC-RPE to facilitate the evaluation of relative pose estimation.
A comprehensive variety of experiments shows that the framework is compatible with various image extractors and matchers, and each of them achieves better performance compared with the variant that relies on the event-to-video pipeline.
We believe our framework provides a new direction for event-image alignment in unconstrained scenarios and could be applied in many multi-modal downstream tasks, like object tracking, SLAM, and visual localization.

%%%%%%%%% REFERENCES
{\small
\bibliographystyle{ieee_fullname}
\bibliography{egbib}

\begin{thebibliography}{10}\itemsep=-1pt

\bibitem{alzugaray2018asynchronous}
Ignacio Alzugaray and Margarita Chli.
\newblock Asynchronous corner detection and tracking for event cameras in real time.
\newblock {\em IEEE Robotics and Automation Letters}, 2018.

\bibitem{balntas2017hpatches}
Vassileios Balntas, Karel Lenc, Andrea Vedaldi, and Krystian Mikolajczyk.
\newblock {HPatches:} {A} benchmark and evaluation of handcrafted and learned local descriptors.
\newblock In {\em CVPR}, 2017.

\bibitem{bellavia2023progressive}
Fabio Bellavia, Luca Morelli, Carlo Colombo, and Fabio Remondino.
\newblock Progressive keypoint localization and refinement in image matching.
\newblock In {\em ICIAPW}, 2023.

\bibitem{burner2022evimo2}
Levi Burner, Anton Mitrokhin, Cornelia Ferm{\"u}ller, and Yiannis Aloimonos.
\newblock {EVIMO2:} {An} event camera dataset for motion segmentation, optical flow, structure from motion, and visual inertial odometry in indoor scenes with monocular or stereo algorithms.
\newblock {\em arXiv preprint arXiv:2205.03467}, 2022.

\bibitem{chen2021learning}
Hongkai Chen, Zixin Luo, Jiahui Zhang, Lei Zhou, Xuyang Bai, Zeyu Hu, Chiew{-}Lan Tai, and Long Quan.
\newblock Learning to match features with seeded graph matching network.
\newblock In {\em ICCV}, 2021.

\bibitem{chen2022efficient}
Jiaan Chen, Hao Shi, Yaozu Ye, Kailun Yang, Lei Sun, and Kaiwei Wang.
\newblock Efficient human pose estimation via {3D} event point cloud.
\newblock In {\em 3DV}, 2022.

\bibitem{chiberre2021detecting}
Philippe Chiberre, Etienne Perot, Amos Sironi, and Vincent Lepetit.
\newblock Detecting stable keypoints from events through image gradient prediction.
\newblock In {\em CVPRW}, 2021.

\bibitem{clady2017motion}
Xavier Clady, Jean-Matthieu Maro, S{\'e}bastien Barr{\'e}, and Ryad~B Benosman.
\newblock A motion-based feature for event-based pattern recognition.
\newblock {\em Frontiers in neuroscience}, 10:594, 2017.

\bibitem{detone2018superpoint}
Daniel DeTone, Tomasz Malisiewicz, and Andrew Rabinovich.
\newblock {SuperPoint:} {Self-supervised} interest point detection and description.
\newblock In {\em CVPRW}, 2018.

\bibitem{ercan2023evreal}
Burak Ercan, Onur Eker, Aykut Erdem, and Erkut Erdem.
\newblock {EVREAL:} {Towards} a comprehensive benchmark and analysis suite for event-based video reconstruction.
\newblock In {\em CVPRW}, 2023.

\bibitem{ercan2024hypere2vid}
Burak Ercan, Onur Eker, Canberk Saglam, Aykut Erdem, and Erkut Erdem.
\newblock {HyperE2VID:} {Improving} event-based video reconstruction via hypernetworks.
\newblock {\em IEEE Transactions on Image Processing}, 2024.

\bibitem{fischler1981random}
Martin~A. Fischler and Robert~C. Bolles.
\newblock Random sample consensus: {A} paradigm for model fitting with applications to image analysis and automated cartography.
\newblock {\em Communications of the ACM}, 1981.

\bibitem{gao2022vector}
Ling Gao, Yuxuan Liang, Jiaqi Yang, Shaoxun Wu, Chenyu Wang, Jiaben Chen, and Laurent Kneip.
\newblock {VECtor:} {A} versatile event-centric benchmark for multi-sensor {SLAM}.
\newblock {\em IEEE Robotics and Automation Letters}, 2022.

\bibitem{gehrig2018asynchronous}
Daniel Gehrig, Henri Rebecq, Guillermo Gallego, and Davide Scaramuzza.
\newblock Asynchronous, photometric feature tracking using events and frames.
\newblock In {\em ECCV}, 2018.

\bibitem{gehrig2020eklt}
Daniel Gehrig, Henri Rebecq, Guillermo Gallego, and Davide Scaramuzza.
\newblock {EKLT:} {Asynchronous} photometric feature tracking using events and frames.
\newblock {\em International Journal of Computer Vision}, 2020.

\bibitem{gehrig2021dsec}
Mathias Gehrig, Willem Aarents, Daniel Gehrig, and Davide Scaramuzza.
\newblock {DSEC:} {A} stereo event camera dataset for driving scenarios.
\newblock {\em IEEE Robotics and Automation Letters}, 2021.

\bibitem{girdhar2023imagebind}
Rohit Girdhar, Alaaeldin El{-}Nouby, Zhuang Liu, Mannat Singh, Kalyan~Vasudev Alwala, Armand Joulin, and Ishan Misra.
\newblock {ImageBind:} {One} embedding space to bind them all.
\newblock In {\em CVPR}, 2023.

\bibitem{gleize2023silk}
Pierre Gleize, Weiyao Wang, and Matt Feiszli.
\newblock {SiLK:} {Simple} learned keypoints.
\newblock In {\em ICCV}, 2023.

\bibitem{harris1988combined}
Christopher~G. Harris and Mike Stephens.
\newblock A combined corner and edge detector.
\newblock In {\em AVC}, 1988.

\bibitem{huang2023eventpoint}
Ze Huang, Li Sun, Cheng Zhao, Song Li, and Songzhi Su.
\newblock {EventPoint:} {Self-supervised} interest point detection and description for event-based camera.
\newblock In {\em WACV}, 2023.

\bibitem{jiao2023lce}
Jianhao Jiao, Feiyi Chen, Hexiang Wei, Jin Wu, and Ming Liu.
\newblock {LCE-Calib:} {Automatic} {LiDAR-frame/event} camera extrinsic calibration with a globally optimal solution.
\newblock {\em IEEE/ASME Transactions on Mechatronics}, 2023.

\bibitem{lagorce2015spatiotemporal}
Xavier Lagorce, Sio-Hoi Ieng, Xavier Clady, Michael Pfeiffer, and Ryad~B Benosman.
\newblock Spatiotemporal features for asynchronous event-based data.
\newblock {\em Frontiers in neuroscience}, 9:46, 2015.

\bibitem{lagorce2014asynchronous}
Xavier Lagorce, C{\'e}dric Meyer, Sio-Hoi Ieng, David Filliat, and Ryad Benosman.
\newblock Asynchronous event-based multikernel algorithm for high-speed visual features tracking.
\newblock {\em IEEE Transactions on Neural Networks and Learning Systems}, 2015.

\bibitem{li2019fa}
Ruoxiang Li, Dianxi Shi, Yongjun Zhang, Kaiyue Li, and Ruihao Li.
\newblock Fa-harris: A fast and asynchronous corner detector for event cameras.
\newblock In {\em 2019 IEEE/RSJ International Conference on Intelligent Robots and Systems (IROS)}, pages 6223--6229. IEEE, 2019.

\bibitem{li2018megadepth}
Zhengqi Li and Noah Snavely.
\newblock {MegaDepth:} {Learning} single-view depth prediction from internet photos.
\newblock In {\em CVPR}, 2018.

\bibitem{lindenberger2023lightglue}
Philipp Lindenberger, Paul-Edouard Sarlin, and Marc Pollefeys.
\newblock {LightGlue:} {Local} feature matching at light speed.
\newblock In {\em ICCV}, 2023.

\bibitem{lowe2004distinctive}
David~G. Lowe.
\newblock Distinctive image features from scale-invariant keypoints.
\newblock {\em International Journal of Computer Vision}, 2004.

\bibitem{manderscheid2019speed}
Jacques Manderscheid, Amos Sironi, Nicolas Bourdis, Davide Migliore, and Vincent Lepetit.
\newblock Speed invariant time surface for learning to detect corner points with event-based cameras.
\newblock In {\em CVPR}, 2019.

\bibitem{messikommer2023data}
Nico Messikommer, Carter Fang, Mathias Gehrig, and Davide Scaramuzza.
\newblock Data-driven feature tracking for event cameras.
\newblock In {\em CVPR}, 2023.

\bibitem{mueggler2017fast}
Elias Mueggler, Chiara Bartolozzi, and Davide Scaramuzza.
\newblock Fast event-based corner detection.
\newblock In {\em BMVC}, 2017.

\bibitem{mueggler2017event}
Elias Mueggler, Henri Rebecq, Guillermo Gallego, Tobi Delbruck, and Davide Scaramuzza.
\newblock The event-camera dataset and simulator: Event-based data for pose estimation, visual odometry, and {SLAM}.
\newblock {\em The International Journal of Robotics Research}, 2017.

\bibitem{muglikar2021calibrate}
Manasi Muglikar, Mathias Gehrig, Daniel Gehrig, and Davide Scaramuzza.
\newblock How to calibrate your event camera.
\newblock In {\em CVPRW}, 2021.

\bibitem{mur2015orb}
Raul Mur{-}Artal, J.~M.~M. Montiel, and Juan~D. Tard{\'{o}}s.
\newblock {ORB-SLAM:} {A} versatile and accurate monocular {SLAM} system.
\newblock {\em IEEE Transactions on Robotics}, 2015.

\bibitem{ramesh2019dart}
Bharath Ramesh, Hong Yang, Garrick Orchard, Ngoc~Anh Le~Thi, Shihao Zhang, and Cheng Xiang.
\newblock {DART:} {Distribution} aware retinal transform for event-based cameras.
\newblock {\em IEEE Transactions on Pattern Analysis and Machine Intelligence}, 2020.

\bibitem{rublee2011orb}
Ethan Rublee, Vincent Rabaud, Kurt Konolige, and Gary Bradski.
\newblock {ORB:} {An} efficient alternative to {SIFT} or {SURF}.
\newblock In {\em ICCV}, 2011.

\bibitem{sarlin2019coarse}
Paul-Edouard Sarlin, Cesar Cadena, Roland Siegwart, and Marcin Dymczyk.
\newblock From coarse to fine: Robust hierarchical localization at large scale.
\newblock In {\em CVPR}, 2019.

\bibitem{sarlin2020superglue}
Paul-Edouard Sarlin, Daniel DeTone, Tomasz Malisiewicz, and Andrew Rabinovich.
\newblock {SuperGlue:} {Learning} feature matching with graph neural networks.
\newblock In {\em CVPR}, 2020.

\bibitem{scheerlinck2020fast}
Cedric Scheerlinck, Henri Rebecq, Daniel Gehrig, Nick Barnes, Robert Mahony, and Davide Scaramuzza.
\newblock Fast image reconstruction with an event camera.
\newblock In {\em WACV}, 2020.

\bibitem{schonberger2016structure}
Johannes~L. Sch{\"{o}}nberger and Jan{-}Michael Frahm.
\newblock Structure-from-motion revisited.
\newblock In {\em CVPR}, 2016.

\bibitem{shi2022csflow}
Hao Shi, Yifan Zhou, Kailun Yang, Xiaoting Yin, and Kaiwei Wang.
\newblock {CSFlow:} {Learning} optical flow via cross strip correlation for autonomous driving.
\newblock In {\em IV}, 2022.

\bibitem{shi2023panoflow}
Hao Shi, Yifan Zhou, Kailun Yang, Xiaoting Yin, Ze Wang, Yaozu Ye, Zhe Yin, Shi Meng, Peng Li, and Kaiwei Wang.
\newblock {PanoFlow:} {Learning} 360$\degree$ optical flow for surrounding temporal understanding.
\newblock {\em IEEE Transactions on Intelligent Transportation Systems}, 2023.

\bibitem{shi2023panovpr}
Ze Shi, Hao Shi, Kailun Yang, Zhe Yin, Yining Lin, and Kaiwei Wang.
\newblock {PanoVPR:} {Towards} unified perspective-to-equirectangular visual place recognition via sliding windows across the panoramic view.
\newblock In {\em ITSC}, 2023.

\bibitem{stoffregen2020reducing}
Timo Stoffregen, Cedric Scheerlinck, Davide Scaramuzza, Tom Drummond, Nick Barnes, Lindsay Kleeman, and Robert~E. Mahony.
\newblock Reducing the sim-to-real gap for event cameras.
\newblock In {\em ECCV}, 2020.

\bibitem{sun2021loftr}
Jiaming Sun, Zehong Shen, Yuang Wang, Hujun Bao, and Xiaowei Zhou.
\newblock {LoFTR:} {Detector-free} local feature matching with transformers.
\newblock In {\em CVPR}, 2021.

\bibitem{sun2022event}
Lei Sun, Christos Sakaridis, Jingyun Liang, Qi Jiang, Kailun Yang, Peng Sun, Yaozu Ye, Kaiwei Wang, and Luc~Van Gool.
\newblock Event-based fusion for motion deblurring with cross-modal attention.
\newblock In {\em ECCV}, 2022.

\bibitem{trajkovic1998fast}
Miroslav Trajkovi{\'c} and Mark Hedley.
\newblock Fast corner detection.
\newblock {\em Image and Vision Computing}, 1998.

\bibitem{tyszkiewicz2020disk}
Micha{\l} Tyszkiewicz, Pascal Fua, and Eduard Trulls.
\newblock {DISK:} {Learning} local features with policy gradient.
\newblock In {\em NeurIPS}, 2020.

\bibitem{vasco2016fast}
Valentina Vasco, Arren Glover, and Chiara Bartolozzi.
\newblock Fast event-based harris corner detection exploiting the advantages of event-driven cameras.
\newblock In {\em IROS}, 2016.

\bibitem{wang2019event}
Lin Wang, S.~Mohammad~Mostafavi I., Yo{-}Sung Ho, and Kuk{-}Jin Yoon.
\newblock Event-based high dynamic range image and very high frame rate video generation using conditional generative adversarial networks.
\newblock In {\em CVPR}, 2019.

\bibitem{wang2022matchformer}
Qing Wang, Jiaming Zhang, Kailun Yang, Kunyu Peng, and Rainer Stiefelhagen.
\newblock {MatchFormer:} {Interleaving} attention in transformers for feature matching.
\newblock In {\em ACCV}, 2022.

\bibitem{wang2023lf}
Ze Wang, Kailun Yang, Hao Shi, Peng Li, Fei Gao, Jian Bai, and Kaiwei Wang.
\newblock {LF-VISLAM:} {A} {SLAM} framework for large field-of-view cameras with negative imaging plane on mobile agents.
\newblock {\em IEEE Transactions on Automation Science and Engineering}, 2023.

\bibitem{wang2024lf}
Ze Wang, Kailun Yang, Hao Shi, Yufan Zhang, Zhijie Xu, Fei Gao, and Kaiwei Wang.
\newblock {LF-PGVIO:} {A} visual-inertial-odometry framework for large field-of-view cameras using points and geodesic segments.
\newblock {\em IEEE Transactions on Intelligent Vehicles}, 2024.

\bibitem{xu2020cross}
Xing Xu, Tan Wang, Yang Yang, Lin Zuo, Fumin Shen, and Heng~Tao Shen.
\newblock Cross-modal attention with semantic consistence for image--text matching.
\newblock {\em IEEE Transactions on Neural Networks and Learning Systems}, 2020.

\bibitem{ye2023towards}
Yaozu Ye, Hao Shi, Kailun Yang, Ze Wang, Xiaoting Yin, Yaonan Wang, and Kaiwei Wang.
\newblock Towards anytime optical flow estimation with event cameras.
\newblock {\em arXiv preprint arXiv:2307.05033}, 2023.

\bibitem{yi2018learning}
Kwang~Moo Yi, Eduard Trulls, Yuki Ono, Vincent Lepetit, Mathieu Salzmann, and Pascal Fua.
\newblock Learning to find good correspondences.
\newblock In {\em CVPR}, 2018.

\bibitem{yi2023focusflow}
Zhonghua Yi, Hao Shi, Kailun Yang, Qi Jiang, Yaozu Ye, Ze Wang, Huajian Ni, and Kaiwei Wang.
\newblock {FocusFlow:} {Boosting} key-points optical flow estimation for autonomous driving.
\newblock {\em IEEE Transactions on Intelligent Vehicles}, 2024.

\bibitem{yin2023rethinking}
Xiaoting Yin, Hao Shi, Jiaan Chen, Ze Wang, Yaozu Ye, Huajian Ni, Kailun Yang, and Kaiwei Wang.
\newblock Rethinking event-based human pose estimation with {3D} event representations.
\newblock {\em Computer Vision and Image Understanding}, 2024.

\bibitem{zhou2020temporal}
Chunluan Zhou, Zhou Ren, and Gang Hua.
\newblock Temporal keypoint matching and refinement network for pose estimation and tracking.
\newblock In {\em ECCV}, 2020.

\bibitem{zhu2018mvsec}
Alex~Zihao Zhu, Dinesh Thakur, Tolga {\"{O}}zaslan, Bernd Pfrommer, Vijay Kumar, and Kostas Daniilidis.
\newblock The multivehicle stereo event camera dataset: An event camera dataset for {3D} perception.
\newblock {\em IEEE Robotics and Automation Letters}, 2018.

\bibitem{zhu2019unsupervised}
Alex~Zihao Zhu, Liangzhe Yuan, Kenneth Chaney, and Kostas Daniilidis.
\newblock Unsupervised event-based learning of optical flow, depth, and egomotion.
\newblock In {\em CVPR}, 2019.

\end{thebibliography}
}

\clearpage

\appendix
\section{Maths}
\subsection{Repeatability}

Given two keypoint sets $E_p=\{ (p_i^{E_p}, d_i^{E_p}) \}$, $I_p=\{ (p_i^{I_p}, d_i^{I_p}) \}$ and the homography $\mathbf{H}$ between two imaging plane, only those which can find a corresponding point within a spatial distance threshold of $\epsilon$ in another set of keypoints after been warped with $\mathbf{H}$ are treated as valid keypoints.
In practice, the $\mathbf{H}$ is an identity matrix when evaluating on events and the corresponding image at the same timestamp.
After that, two valid keypoint sets $E_{valid}$ and $I_{valid}$ are filtered out, and $\{(p_i^{E_{valid}}, d_i^{E_{valid}}), (p_i^{I_{valid}}, d_i^{I_{valid}}) \}$ are the corresponding valid keypoints.
Then, the \textit{Repeatability} is computed as following:
\begin{equation}
    \textit{Repeatability} = \frac{|E_{valid}|+|I_{valid}|}{|E_{p}|+|I_{p}|}.
\end{equation}

\subsection{VDD and VDA}
Given the valid keypoint sets $E_{valid}$ and $I_{valid}$, the \textit{valid descriptor distance (VDD)} and \textit{valid distance angle (VDA)} are computed as following:
\begin{equation}
    \textit{VDD} = \frac{1}{N}\sum_{i=1}^N{\lVert d_i^{E_{valid}}-d_i^{I_{valid}} \rVert _2},
\end{equation}
\begin{equation}
    \textit{VDA} = \frac{1}{N}\sum_{i=1}^N{\arccos(\lVert d_i^{E_{valid}}-d_i^{I_{valid}} \rVert _2)}.
\end{equation}

\subsection{RPE Ratio and RPE AUC}
Given two matched keypoint sets, the essential matrix $\mathbf{E}$ is firstly estimated by using \textit{cv.findEssentialMat()} with RANSAC, then the estimated rotation $\mathbf{\hat{R}}$ and translation $\mathbf{\hat{t}}$ are then recovered.
When a list of relative pose estimation results $\{ \mathbf{\hat{R}}, \mathbf{\hat{t}} \}$ with corresponding ground truth $\{ \mathbf{R}^{gt}, \mathbf{t}^{gt} \}$, the error of $\mathbf{\hat{R}}$ and $\mathbf{\hat{t}}$ are defined as the angular error between the estimation and the ground truth:
\begin{equation}
\begin{split}
    \mathbf{R}^{err} =& \frac{tr \left( \mathbf{\hat{R}}^{\top}\mathbf{R}^{gt} \right) -1}{2},\\
    \mathbf{t}^{err} =& \frac{\mathbf{\hat{t}}-\mathbf{t}^{gt}}{\lVert \mathbf{\hat{t}} \rVert \cdot \lVert \mathbf{t}^{gt} \rVert}.
\end{split}
\end{equation}
The pose error $err_i$ of the current estimation is defined as the maximum error of $\mathbf{R}^{err}_i$ and $\mathbf{t}^{err}_i$.
The \textit{RPE Ratio} is then computed:
\begin{equation}
    \textit{Ratio} = \frac{|\{err_i \le \epsilon \}|}{|\{ err_i \}|},
\end{equation}
where $\epsilon$ is the specified threshold of angle.
For RPE AUC, the area under curve (AUC) is calculated following SiLK~\cite{gleize2023silk}.
Given a threshold $\epsilon$ and the estimation errors $\{err_i\}$, the calculation code is described in Algorithm~\ref{alg:rpe auc}.

%##################################################################################################
\begin{algorithm}[t]
\caption{RPE AUC Pseudocode, Numpy-like}
\label{alg:rpe auc}
\definecolor{codeblue}{rgb}{0.25,0.5,0.5}
\definecolor{codekw}{rgb}{0.85, 0.18, 0.50}
\lstset{
  backgroundcolor=\color{white},
  basicstyle=\fontsize{7.5pt}{7.5pt}\ttfamily\selectfont,
  columns=fullflexible,
  breaklines=true,
  captionpos=b,
  commentstyle=\fontsize{7.5pt}{7.5pt}\color{codeblue},
  keywordstyle=\fontsize{7.5pt}{7.5pt}\color{codekw},
}
\begin{lstlisting}[language=python]
def compute_auc(errors, threshold):
    errors = errors[np.isfinite(errors)]

    sort_idx = np.argsort(errors)
    errors = np.array(errors.copy())[sort_idx]
    recall = (np.arange(len(errors)) + 1) / len(errors)
    errors = np.r_[0.0, errors]
    recall = np.r_[0.0, recall]

    last_index = np.searchsorted(errors, threshold)
    rec = np.r_[recall[:last_index], recall[last_index - 1]]
    err = np.r_[errors[:last_index], threshold]
    auc = np.trapz(rec, x=err) / threshold
    return auc
\end{lstlisting}
\end{algorithm}
%##################################################################################################

\subsection{Groundtruth assignment}
Given two keypoint sets $E_p$ and $I_p$, corresponding depth map $d^E$ and $d^I$, camera matrix $\mathbf{K}^E$ and $\mathbf{K}^I$, and current pose $\mathbf{R}^E, \mathbf{t}^E, \mathbf{R}^I, \mathbf{t}^I$ of event view and image view, the relative poses $\mathbf{T}^{E\rightarrow I}$ and $\mathbf{T}^{I\rightarrow E}$ are computed first.
Then, the keypoints of one view are projected into the other view:
\begin{equation}
\begin{split}
    \mathbf{p}_{i}^{E\rightarrow I}=&\mathbf{K}^I\frac{1}{z}\mathbf{T}^{E\rightarrow I}d_{i}^{E}\left( \mathbf{K}^E \right) ^{-1}\left[ \mathbf{p}_{i}^{E},1 \right] ^{\top},\\
    \mathbf{p}_{j}^{I\rightarrow E}=&\mathbf{K}^E\frac{1}{z}\mathbf{T}^{I\rightarrow E}d_{j}^{I}\left( \mathbf{K}^I \right) ^{-1}\left[ \mathbf{p}_{j}^{I},1 \right] ^{\top},
\end{split}
\end{equation}
where $z$ is the normalization term to normalize the depth of the points into unit length.
Then the re-projection distance matrix $\mathbf{D}^{M\times N}$ is computed:
\begin{equation}
    \mathbf{D}_{ij}=\max \left( \lVert \mathbf{p}_{i}^{E\rightarrow I}-\mathbf{p}_{j}^{I} \rVert_2^2 ,\lVert \mathbf{p}_{i}^{E}-\mathbf{p}_{j}^{I\rightarrow E} \rVert_2^2 \right).
\end{equation}
According to $D^{M\times N}$, the ground-truth assignment $\mathbf{P}^{M\times N}$ is then calculated.
Elements $\mathbf{P}_{ij}$ are marked as positive, only if they have the minimum distance in both $i$-th row and $j$-th column, and the distance $\mathbf{D}_{ij}$ smaller than a threshold $\epsilon_p ^2$:
\begin{equation}
\mathbf{P}_{ij}=\left\{ \begin{array}{l}
	1,\mathbf{D}_{ij}\le \mathbf{D}_{:j},\mathbf{D}_{ij}\le \mathbf{D}_{i:},\mathbf{D}_{ij}<\varepsilon _{p}^{2};\\
	0,others.\\
\end{array} \right. 
\end{equation}
Finally, the ground-truth matches $\text{M}^{gt}$ are obtained through selecting all the $(i,j)$ pairs that have $\mathbf{P}_{ij}{=}1$.

\begin{table*}[t]
\centering
\renewcommand{\arraystretch}{1.1}
\setlength{\tabcolsep}{10pt}
\resizebox{\linewidth}{!}{

\begin{tabular}{c|lcccccccccc}
\hline
\multirow{2}{*}{\begin{tabular}[c]{@{}c@{}}\textbf{Image}\\ \textbf{Extractor}\end{tabular}} & \multicolumn{1}{c}{\multirow{2}{*}{\textbf{Method}}} & \multicolumn{2}{c}{\textbf{\underline{\hspace{1cm}MMA\hspace{1cm}}}} & \multirow{2}{*}{\textbf{MR}} & \multirow{2}{*}{\textbf{HE Inlier}} & \multicolumn{3}{c}{\textbf{\underline{\hspace{1cm}HE Ratio\hspace{1cm}}}} & \multicolumn{3}{c}{\textbf{\underline{\hspace{1cm}HE AUC\hspace{1cm}}}} \\
 & \multicolumn{1}{c}{} & $\epsilon=1$ & $\epsilon=3$ &  &  & $\epsilon=3$ & $\epsilon=5$ & $\epsilon=10$ & $\epsilon=3$ & $\epsilon=5$ & $\epsilon=10$ \\ \hline \hline
\multirow{3}{*}{SuperPoint} & E2VID & 0.248 & 0.571 & 0.448 & 0.503 & 0.560 & 0.772 & 0.939 & \textbf{24.17} & 41.53 & 64.55 \\
 & HyperE2VID & 0.209 & 0.511 & 0.395 & 0.468 & 0.434 & 0.691 & 0.898 & 16.22 & 33.62 & 59.01 \\
 & Ours (SuperPoint) & \textbf{0.249} & \textbf{0.660} & \textbf{0.546} & \textbf{0.571} & \textbf{0.585} & \textbf{0.843} & \textbf{0.959} & 21.00 & \textbf{42.90} & \textbf{67.77} \\ \hline
\multirow{3}{*}{SiLK} & E2VID & 0.145 & 0.303 & \textbf{0.331} & 0.279 & 0.626 & 0.772 & 0.838 & 29.32 & 46.27 & 64.09 \\
 & HyperE2VID & 0.097 & 0.217 & 0.317 & 0.201 & 0.444 & 0.611 & 0.762 & 18.58 & 32.88 & 51.89 \\
 & Ours (SiLK) & \textbf{0.267} & \textbf{0.485} & 0.258 & \textbf{0.456} & \textbf{0.691} & \textbf{0.883} & \textbf{0.944} & \textbf{32.23} & \textbf{51.44} & \textbf{72.16} \\ \hline
\end{tabular}
}
\caption{\textbf{MMA, MR and HE results on EC-RPE set.} MNN is employed for feature matching.}
\label{table:supp_mma_he}
\vskip -1.0\baselineskip plus -1fil
\end{table*}

\section{More Implementation Details}
\subsection{Implementation with SuperPoint}
We utilize the SuperPoint architecture and the pre-trained model from the official LightGlue training repository~\cite{lindenberger2023lightglue}, which consists of $1.30M$ parameters in total.
The CNN-based backbone encodes the grayscale image $I^{H\times W}$ into latent feature $I_f^{\frac{H}{8}\times \frac{W}{8}\times 128}$.
Score head and descriptor head separately predict a score map $I_{score}^{\frac{H}{8}\times \frac{W}{8}\times 65}$ and a descriptor map $I_{desc}^{\frac{H}{8}\times \frac{W}{8}\times 256}$.
Then the dustbin dimension of $I_{score}$ is removed and $I_{score}^{\frac{H}{8}\times \frac{W}{8}\times 64}$ is reshaped into $I_{score}^{H\times W\times 1}$ through pixel shuffle.
After that, the keypoint extraction procedure from SiLK~\cite{gleize2023silk} is employed, which produces a set of keypoint positions $p_i$.
The $d_i$ of the keypoint in $p_i$ is then extracted from the normalized $I_{desc}$ through bilinear sampling.

When employing the event extractor $\mathcal{E}_E$ corresponding to SuperPoint, the dimensions of the latent feature, score map, and descriptor map are supposed to be the same as SuperPoint, due to the implementation of the proposed local feature distillation.
In practice, we construct a VGG-like architecture as $\mathcal{E}_E$ that has $1.31M$ parameters.
During training, we perform the cosine learning schedule with an initial learning rate of $1{\times}10^{-3}$ for $50$ epochs, and the batch size is set to $8$.

\subsection{Implementation with SiLK}
We perform the official SiLK model that uses no max-pooling and consists of $1.57M$ parameters.
In this case, it generates $I_f^{H\times H\times 128}$, $I_{score}^{H\times W\times 1}$, $I_{desc}^{H\times W\times 128}$ in full resolution.
Therefore, we do not use max-pooling in our VGG-like $\mathcal{E}_E$ with $1.10M$ parameters when training with SiLK.
The learning hyper-parameters are set the same as used in the SuperPoint implementation.

\subsection{Mutual Nearest Neighbor}
Give two keypoint sets $E_p{=}\left\{ p^E_i,d^E_i \right\} $ and $I_p{=}\left\{ p^I_i,d^I_i \right\} $, a similarity matrix $\mathbf{S}^{M\times N}$ is firstly computed:
\begin{equation}
    \mathbf{S}_{ij}=\left( d_{i}^{E} \right) ^{\top}d_{j}^{I}.
\end{equation}
The estimated assignment $\mathbf{\hat{P}}$ is obtained by applying a softmax operation followed by a logarithm on each axis:
\begin{equation}
    \mathbf{\hat{P}}=\log \left( \frac{\exp \left( \mathbf{S}_{ij} \right)}{\sum{\exp \left( \mathbf{S}_{:j} \right)}} \right) +\log \left( \frac{\exp \left( \mathbf{S}_{ij} \right)}{\sum{\exp \left( \mathbf{S}_{i:} \right)}} \right) .
\end{equation}
The final predicted matches $\hat{\text{M}}$ are obtained through filtering all the possible $(i,j)$ pairs where $\mathbf{\hat{P}}_{ij}$ have the largest value among the $i$-th row and the $j$-th column.

\subsection{LightGlue}
LightGlue is a representative Context Aggregation (CA) method that uses attention techniques to aggregate information within the keypoint set and between keypoint sets.
We follow the official implementation of LightGlue, but ignore the point pruning and early stop procedure, which are designed to boost inference speed and do not affect the matching performance.
During training, we utilize a cosine schedule with an initial learning rate of $1{\times}10^{-4}$ for $50$ epochs.
The batch size is set to $8$.

\section{Additional Experiments}
\subsection{More Cross-modal Keypoint Similarity Results}
\noindent \textbf{Mean Matching Accuracy and Matching Rate}.
Following SiLK~\cite{gleize2023silk}, we further evaluate the cross-modal keypoint similarity with \textit{mean matching accuracy (MMA)}.
Given the events and image at the same timestamp, the MMA measures the accuracy of the valid matching pairs by applying an MNN for feature matching.
We set two thresholds $\epsilon{=}1$ and $\epsilon{=}3$ to obtain valid checks.
In addition, the \textit{matching rate (MR)} which calculates the ratio of the matched pairs is also evaluated for a more comprehensive comparison.
The MMA and MR results are shown in Table~\ref{table:supp_mma_he}.
Our framework surpasses the explicit transform methods by a large margin in most situations, showing the best keypoint similarity among all methods.

\begin{figure*}[t]
    \centering
    \includegraphics[width=\linewidth]{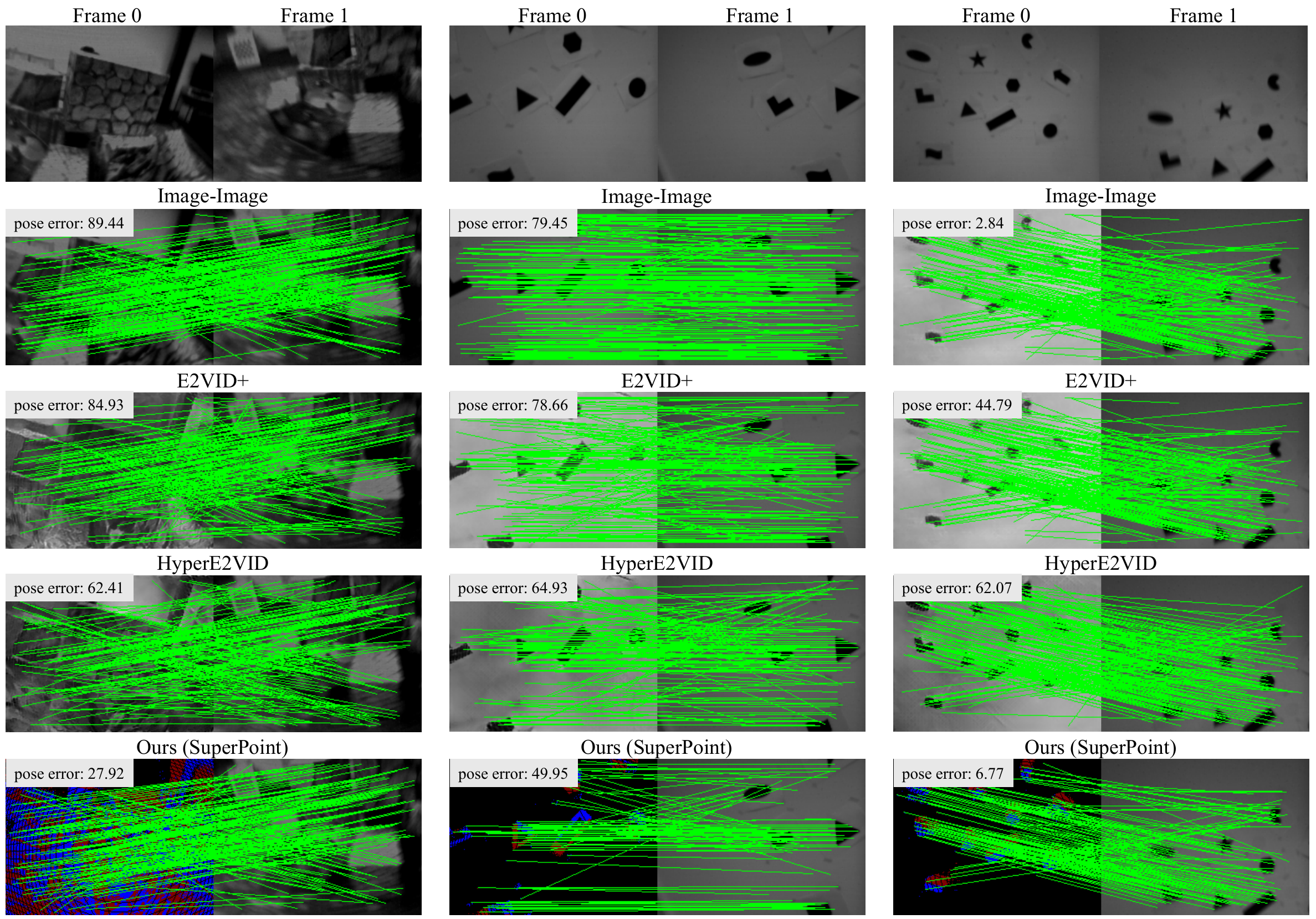}
    \caption{\textbf{Matching results on EC-RPE set for relative pose estimation.} Features from the image are extracted by SuperPoint.}
    \label{fig:supp_ec_rpe}
    \vskip -1.0\baselineskip plus -1fil
\end{figure*}

\noindent \textbf{Homography Estimation}.
Since the events and image are at the same timestamp when evaluating the cross-modal keypoint similarity, the homography between two imaging planes is an identity transform.
We follow previous feature matching methods~\cite{detone2018superpoint, gleize2023silk, sarlin2020superglue} to construct a Homography Estimation (HE) task for event-image feature matching.
The matched keypoint pairs are used for estimating a homography $\mathbf{\hat{H}}$ using \textit{cv.findHomography()} with RANSAC.
The corner error between the images warped with the estimated homography $\mathbf{\hat{H}}$ and the ground-truth homography $\mathbf{H}^{gt}{=}\mathbf{I}$ is then computed for correctness identification.
Lastly, the \textit{HE Inlier} ratio from RANSAC, the \textit{HE Ratio} and \textit{HE AUC} under different thresholds are computed.
As shown in Table~\ref{table:supp_mma_he}, our method achieves the most accurate estimation result, emphasizing the superiority of the direct inter-modality feature matching proposed by EI-Nexus.

\subsection{More Relative Pose Estimation Results}

We further show the inter-modality RPE results on the EC-RPE set.
As shown in Fig.~\ref{fig:supp_ec_rpe}, the RPE results of EI-Nexus are much better than event-to-video methods, and even better than image-image results for some instances.
It indicates that the image local feature extraction methods that trained on a specific dataset, could not always perform well in new scenes, and no suitable fine-tuning approach for now is presented for image-based local extraction.
In contrast, our proposed framework can achieve superior and stable performance through the use of the simple yet effective local feature distillation (LFD) approach.
This highlights the broad applicability and ease of implementation of our method.

\begin{figure*}[ht]
        \resizebox{0.3\linewidth}{!}{
        \begin{tikzpicture}
            \begin{axis}[
                xtick={128, 256, 512, 1024, 2048}, %
                % legend pos=south west,
                xticklabels={128, 256, 512, 1024, 2048}, %
                ymin=0.1,
                grid=both,
                xmode=log,
                grid style={line width=.1pt, draw=gray!10},
                major grid style={line width=.2pt,draw=gray!50},
                minor tick num=2,
                % axis x line*=bottom,
                % axis y line*=left,
                height=0.4\linewidth,
                width=0.5\linewidth,
                ylabel style= {align=center, font=\large},
                xlabel style = {font=\large, font=\large},
                ylabel={Repeatability $\uparrow$ ($\epsilon{=}1$)},
                xlabel={$k$ in top-$k$ selection strategy},
                yticklabel style = {font=\large},
                xticklabel style = {font=\large},
                legend style ={ at={(0.65,0.65)}, anchor=north west, draw=black, fill=white, align=left, font=\normalsize},
            ]
            % NMS=0
            \addplot[mark=o, very thick, magenta, mark options={solid}, line width=1pt, mark size=3pt] plot coordinates {
                (128, 0.418) %
                (256, 0.458) %
                (512, 0.487) %
                (1024, 0.510) %
                (2048, 0.532) %
            };
            \addlegendentry{NMS=0}
            % NMS=4
            \addplot[mark=triangle, very thick, brown, mark options={solid}, line width=1pt, mark size=3pt] plot coordinates {
                (128, 0.2829) %
                (256, 0.2341) %
                (512, 0.2012) %
                (1024, 0.1785) %
                (2048, 0.1809) %
            };
            \addlegendentry{NMS=4}
            % NMS=9
            \addplot[mark=square, very thick, teal, mark options={solid}, line width=1pt, mark size=3pt] plot coordinates {
                (128, 0.2549) %
                (256, 0.1893) %
                (512, 0.1338) %
                (1024, 0.1313) %
                (2048, 0.1313) %
            };
            \addlegendentry{NMS=9}
        
            \end{axis}
    
        \end{tikzpicture}
        }
    \hfill
    \resizebox{0.3\linewidth}{!}{
        \begin{tikzpicture}
            \begin{axis}[
                xtick={128, 256, 512, 1024, 2048}, %
                % legend pos=south west,
                xticklabels={128, 256, 512, 1024, 2048}, %
                ymin=0.15,
                grid=both,
                xmode=log,
                grid style={line width=.1pt, draw=gray!10},
                major grid style={line width=.2pt,draw=gray!50},
                minor tick num=2,
                % axis x line*=bottom,
                % axis y line*=left,
                height=0.4\linewidth,
                width=0.5\linewidth,
                ylabel style= {align=center, font=\large},
                xlabel style = {font=\large, font=\large},
                ylabel={RPE AUC for \YZH{MNN} $\uparrow$ ($\epsilon{=}20$)},
                xlabel={$k$ in top-$k$ selection strategy},
                yticklabel style = {font=\large},
                xticklabel style = {font=\large},
                legend style ={ at={(0.03,0.3)}, anchor=north west, draw=black, fill=white, align=left, font=\normalsize},
            ]
            % NMS=0
            \addplot[mark=o, very thick, magenta, mark options={solid}, line width=1pt, mark size=3pt] plot coordinates {
                (128, 0.2119) %
                (256, 0.2543) %
                (512, 0.2952) %
                (1024, 0.2659) %
                (2048, 0.2673) %
            };
            \addlegendentry{NMS=0}
            % NMS=4
            \addplot[mark=triangle, very thick, brown, mark options={solid}, line width=1pt, mark size=3pt] plot coordinates {
                (128, 0.2988) %
                (256, 0.2880) %
                (512, 0.2565) %
                (1024, 0.2340) %
                (2048, 0.2227) %
            };
            \addlegendentry{NMS=4}
            % NMS=9
            \addplot[mark=square, very thick, teal, mark options={solid}, line width=1pt, mark size=3pt] plot coordinates {
                (128, 0.2714) %
                (256, 0.2140) %
                (512, 0.1726) %
                (1024, 0.1776) %
                (2048, 0.1776) %
            };
            \addlegendentry{NMS=9}
        
            \end{axis}
    
        \end{tikzpicture}
    }
    \hfill
    \resizebox{0.3\linewidth}{!}{
        \begin{tikzpicture}
            \begin{axis}[
                xtick={128, 256, 512, 1024, 2048}, %
                % legend pos=south west,
                xticklabels={128, 256, 512, 1024, 2048}, %
                ymin=0.15,
                grid=both,
                xmode=log,
                grid style={line width=.1pt, draw=gray!10},
                major grid style={line width=.2pt,draw=gray!50},
                minor tick num=2,
                % axis x line*=bottom,
                % axis y line*=left,
                height=0.4\linewidth,
                width=0.5\linewidth,
                ylabel style= {align=center, font=\large},
                xlabel style = {font=\large, font=\large},
                ylabel={RPE AUC for \YZH{LG} $\uparrow$ ($\epsilon{=}20$)},
                xlabel={$k$ in top-$k$ selection strategy},
                yticklabel style = {font=\large},
                xticklabel style = {font=\large},
                legend style ={ at={(0.65,0.3)}, anchor=north west, draw=black, fill=white, align=left, font=\normalsize},
            ]
            % NMS=0
            \addplot[mark=o, very thick, magenta, mark options={solid}, line width=1pt, mark size=3pt] plot coordinates {
                (128, 0.2930) %
                (256, 0.3589) %
                (512, 0.4396) %
                (1024, 0.4983) %
                (2048, 0.5505) %
            };
            \addlegendentry{NMS=0}
            % NMS=4
            \addplot[mark=triangle, very thick, brown, mark options={solid}, line width=1pt, mark size=3pt] plot coordinates {
                (128, 0.4711) %
                (256, 0.5652) %
                (512, 0.6720) %
                (1024, 0.7273) %
                (2048, 0.7232) %
            };
            \addlegendentry{NMS=4}
            % NMS=9
            \addplot[mark=square, very thick, teal, mark options={solid}, line width=1pt, mark size=3pt] plot coordinates {
                (128, 0.4947) %
                (256, 0.6527) %
                (512, 0.7053) %
                (1024, 0.6970) %
                (2048, 0.6970) %
            };
            \addlegendentry{NMS=9}
        
            \end{axis}
    
        \end{tikzpicture}
    }
    \vskip -0.6\baselineskip plus -1fil
    \caption{\textbf{Comparison of different post-processing parameters of keypoint extraction and different matchers.} The keypoint similarity and RPE metrics are evaluated and compared across a variety of different test scenarios.}
    \label{fig:NMS-topk}
    \vskip -1.0\baselineskip plus -1fil
\end{figure*}
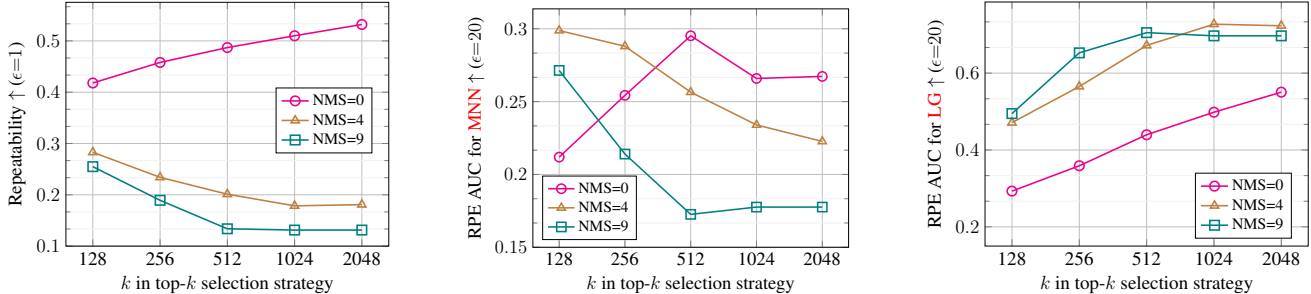

\subsection{Extraction Post-processing Parameters and Properties of Matchers.}
The hyper-parameters governing the keypoint extraction process exert a decisive influence on the quality of the final keypoints.
% In general, these parameters are configured to facilitate model adaptation to diverse scenarios. 
% However, they can also impact the results within the same scene when employing varying parameter combinations.
Since the border removal is to prevent keypoint selection from fake edge information, we do not consider that and investigate the influence of the NMS radius and top-$k$ instead.

As presented in Fig.~\ref{fig:NMS-topk}, it is observed that the trends for the settings of $NMS{=}4$ and $NMS{=}9$ exhibit similar patterns.
When enlarging $k$, the \textit{Repeatability} decreases, meaning the quality of the keypoint set declines.
Concurrently, the MNN exhibits worse RPE performance when $k$ is larger, while the LG behaves in the opposite manner.
We also notice that the performance with $NMS{=}9$ does not change when the $k$ value exceeds $1024$, as the top-$k$ keypoint selection is unable to detect additional keypoints when the score map becomes too sparse after applying a large NMS radius.
In addition, the performance using an NMS radius of $4$ is usually better than $9$, except for the \emph{RPE AUC} when utilizing the LG approach. This is because the usage of the position encoding guides the LG method to focus more on the points that are spread widely.

When NMS is not applied, the extracted keypoints are clustered since the scores around a keypoint are usually close to it.
In this case, a region of keypoints will be extracted without NMS, resulting in a higher probability of having an intersection area with another region.
However, the matching results are not satisfying when applying MNN because lack of points in different fields of view, unless the extracted points are enough but not too much.
In addition, for the LG method, the \textit{RPE AUC} could not achieve good results because of the use of position encoding. 

% Simple
\begin{table}[t]
\large
\centering
\renewcommand{\arraystretch}{1.1}
\resizebox{1.0\columnwidth}{!}{
\setlength{\tabcolsep}{6pt}

\begin{tabular}{l|ccccccc}
\hline
\multirow{2}{*}{\textbf{Method}} & \multirow{2}{*}{\textbf{AVG Pose Error$\downarrow$}} & \multicolumn{3}{c}{\underline{\hspace{1cm}\textbf{RPE Ratio} $\uparrow$\hspace{1cm}}} & \multicolumn{3}{c}{\underline{\hspace{1cm}\textbf{RPE AUC} $\uparrow$\hspace{1cm}}} \\
 &  & $\epsilon=5\degree$ & $\epsilon=10\degree$ & $\epsilon=20\degree$ & $\epsilon=5\degree$ & $\epsilon=10\degree$ & $\epsilon=20\degree$ \\ \hline \hline
HyperE2VID & 48.46 & 0.00 & 0.17 & 0.21 & 0.00 & 7.04 & 13.54 \\
\textbf{Ours} & \textbf{42.32} & \textbf{0.04} & \textbf{0.26} & \textbf{0.34} & \textbf{2.74} & \textbf{10.06} & \textbf{21.03} \\\hline
\end{tabular}
}
\vskip -0.3\baselineskip plus -1fil
\caption{\textbf{Relative pose estimation results on the EVIMO2 dataset.}}
\label{table:evimo2}
\vskip -1.0\baselineskip plus -1fil
\end{table}

\begin{figure}[ht]
    \centering
    \includegraphics[width=\linewidth]{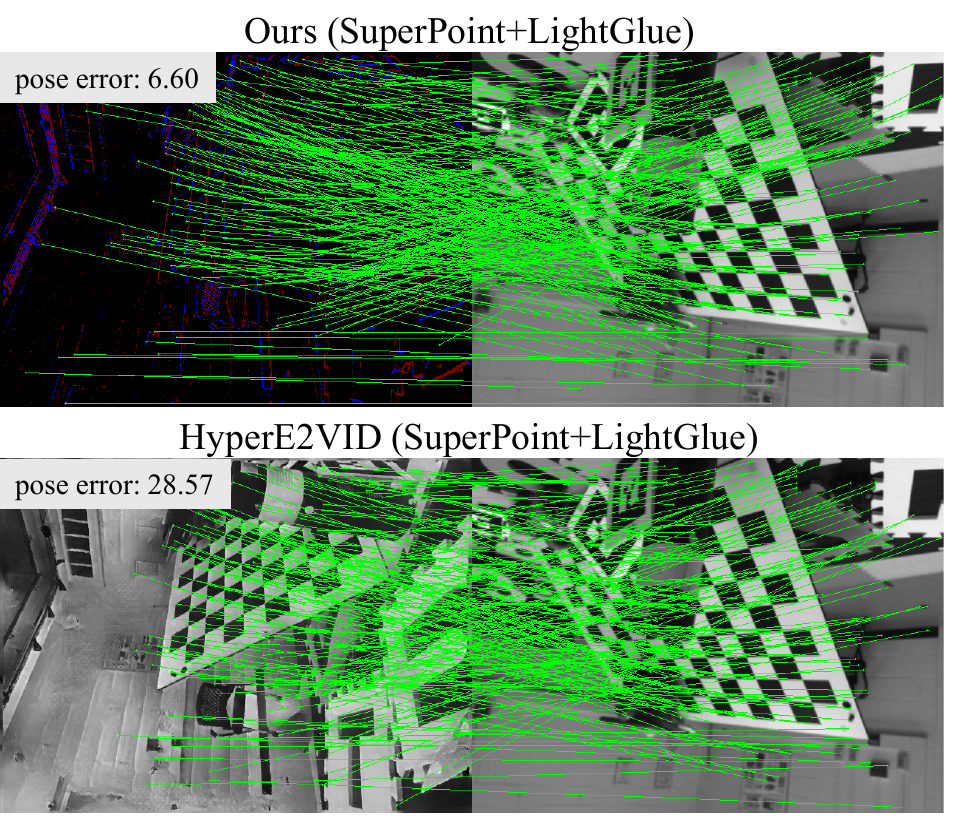}
    \caption{\textbf{Matching results on EVIMO2 dataset.} Features are extracted by SuperPoint and matched by LightGlue.}
    \label{fig:supp_evimo2}
    \vskip -1.0\baselineskip plus -1fil
\end{figure}

The observations presented above suggest that the post-processing procedure of keypoint extraction is highly important for local feature extraction and downstream feature-matching tasks. 
In addition, the analyses underscore the importance of choosing appropriate post-processing parameters for specific scenarios and models.

\subsection{Results on Depth-aware System}

For depth-aware systems~\cite{gehrig2021dsec, gao2022vector, burner2022evimo2}, in which the event camera and regular camera are deployed separately, the pixel-level correspondence could be obtained by reprojecting the image from the imaging plane of the regular camera into the imaging plane of event camera according to the extrinsic parameters and the depth of each pixel in the original image.

Following this pipeline, we test our framework on the EVIMO2 dataset, which contains an RGB camera with $2080{\times}1552$ resolution and two Prophesee cameras with $640{\times}480$ resolution.
We use the sequences from the \textit{sfm} scenario in our experiments.

The event extractor is trained through LFD using the events and reprojected images.
Then we train LightGlue as a learnable matcher using the events and original images, given their ground-truth relative poses.
For evaluation, the events from the event camera and the original image from the regular camera at the same timestamp are used for matching.
The matches are then used to estimate the extrinsic parameter between those two cameras.
Results are shown in Table~\ref{table:evimo2} and quantitative results are shown in Fig.~\ref{fig:supp_evimo2}.
In such a depth-aware system, our model still works better than those that apply explicit modality transformation first.
It should be emphasized that our method only needs events from a short time interval $[t_j-\Delta t, t_j]$, while the event-to-video methods need long-term previous information of the sequence.

\subsection{Inference Time}

\begin{table}[t]
\Large
\centering
\renewcommand{\arraystretch}{1.1}
\resizebox{\linewidth}{!}{

\begin{tabular}{l|c|lcc}
\hline
Method & Data Processing (ms) & Model & Extractor (ms) & Matcher (ms) \\ \hline \hline
\multirow{2}{*}{E2VID+} & \multirow{2}{*}{10.9} & SP+MNN & 18.5 & 35.2 \\
 &  & SP+LG & 17.8 & 51.7 \\ \cline{1-2}
\multirow{2}{*}{HyperE2VID} & \multirow{2}{*}{10.8} & SiLK+MNN & 29.1 & 37.7 \\
 &  & SiLK+LG & 29.3 & 49.1 \\ \hline \hline
\multirow{4}{*}{Ours} & \multirow{4}{*}{46.0} & SP+MNN & 19.1 & 36.5 \\
 &  & SP+LG & 19.1 & 53.5 \\
 &  & SiLK+MNN & 29.3 & 34.5 \\
 &  & SiLK+LG & 28.9 & 54.5 \\ \hline
\end{tabular}
}
\caption{\textbf{Comparison on inference time.} SP represents SuperPoint~\cite{detone2018superpoint} and LG represents LightGlue~\cite{lindenberger2023lightglue}.}
\label{table:inference_time}
% \vskip -0.8\baselineskip plus -1fil
\end{table}

We give the inference time of the models tested on an NVIDIA A800 GPU.
Data processing time for event-to-video methods means the time consumption for reconstructing one frame.
Our model converts the voxel grid and computes an event mask during data processing.
As shown in Table~\ref{table:inference_time}, the entire time consumption of our model lies in the data processing procedure. For the computation cost of the network, the inference time is almost the same as event-to-video methods, as the only difference of the network between ours and events-to-video methods is the input channels of the first CNN layer.
In addition, we find that the SiLK costs more computation than SuperPoint as its backbone does not have a pooling operation.

\section{Limitations}
Despite the impressive performance of EI-Nexus in event-image inter-modality local feature extraction and matching, there remain several limitations that warrant further investigation.
First, The event representations explored in this work are the only commonly used methods that convert the event stream into 2D representations, which may not fully capture the spatial-temporal information for inter-modality local feature extraction and matching.
Future research could explore more expressive representation approaches, such as learning-based techniques, to improve the robustness of the framework.
Second, the intra-modality performance of the learned event extractor is not investigated.
Although the whole EI-Nexus framework is designed for inter-modality tasks, a unified framework for both intra-modality and inter-modality local feature extraction and matching is preferred in downstream applications.
However, since EI-Nexus provides a simple, direct, and flexible approach to constructing the local feature relationship of different modalities, it could be an indispensable part of such a unified multi-modality framework.
Third, the proposed EI-Nexus solution is supposed to train on a specific dataset, which limits its generalization ability.
In this case, further research on training on synthetic data is encouraged.

\end{document}